\documentclass[11pt]{article}

\usepackage{xspace}
\usepackage{enumerate}
\usepackage{graphicx}
\usepackage{booktabs}
\usepackage{xcolor}
\usepackage{pifont}
\usepackage{pagecolor}
\usepackage{titlesec}
\usepackage{caption}
\usepackage{subcaption}
\usepackage[table]{xcolor}
\usepackage[most]{tcolorbox}
\tcbuselibrary{listings, skins, breakable}
\usepackage{listings}
\usepackage[percent]{overpic} 
\usepackage{hyperref}
\usepackage[capitalize]{cleveref}
\usepackage{tikz}

\crefname{section}{Sec.}{Secs.}
\Crefname{section}{Section}{Sections}
\Crefname{table}{Table}{Tables}
\crefname{table}{Tab.}{Tabs.}

\newcommand{\xmark}{\ding{55}}

\usepackage[most]{tcolorbox}
\tcbuselibrary{listings, skins, breakable}
\usepackage{xcolor}
\usepackage{listings}

\definecolor{neoblue}{RGB}{179,217,242}
\definecolor{codegray}{gray}{0.15}
\definecolor{coderule}{RGB}{150,200,230}

\lstdefinestyle{neopy}{
  language=Python,
  basicstyle=\ttfamily\small,
  keywordstyle=\color{blue!70!black}\bfseries,
  stringstyle=\color{green!40!black},
  commentstyle=\color{black!60},
  numberstyle=\tiny\color{black!50},
  numbers=left,
  numbersep=8pt,
  showstringspaces=false,
  breaklines=true,
  tabsize=4,
  keepspaces=true,
}

\newtcblisting{neocodelst}[1]{
  listing only,
  listing options={style=neopy},
  colback=white,
  colframe=neoblue,
  enhanced,
  sharp corners,
  boxrule=0.8pt,
  left=8pt, right=8pt, top=6pt, bottom=6pt,
  breakable,
  title={#1},
  attach boxed title to top left = {yshift=-2mm, xshift=2mm},
  boxed title style = {colback=white, colframe=neoblue, boxrule=0.8pt},
  borderline west  = {3pt}{0pt}{neoblue!90},
}

\definecolor{tabbaseline}{rgb}{0.7, 0.85, 0.95} 
\definecolor{tabfirst}{rgb}{1, 0.7, 0.7} 
\definecolor{tabsecond}{rgb}{1, 0.85, 0.7} 
\definecolor{tabthird}{rgb}{1, 1, 0.7} 

\definecolor{rowblue}{RGB}{220,230,240}
\definecolor{myorchid}{RGB}{150,10,30}
\definecolor{myblue}{RGB}{10,30,250}
\definecolor{mygreen}{RGB}{10,120,10}

\newcommand{\name}{Neodragon\xspace}

\newcommand{\tabcapsfx}{We color code all the scores as \colorbox{tabbaseline}{\textbf{baseline}}, \colorbox{tabfirst}{\textbf{best}}, \colorbox{tabsecond}{\textbf{second best}}, and \colorbox{tabthird}{\textbf{third best}}\xspace}

\newcommand{\platform}{Qualcomm Hexagon NPU}
\newcommand{\laptopsoc}{Snapdragon X Elite}
\newcommand{\mobilesoc}{Snapdragon 8 Elite Gen4}

\newcommand{\redcross}{\textcolor{red!70!black}{\xmark}}
\newcommand{\greentick}{\textcolor{mygreen}{\textbf{\checkmark}}}
\newcommand{\antichrist}{\rotatebox[origin=c]{180}{\textdagger}}

\newcommand{\baseline}{Pyramidal-Flow\xspace}

\newcommand{\tfxxl}{$\mathit{T5}_\text{XXL}$\xspace}
\newcommand{\dtf}{$\mathit{DT5}$\xspace}
\newcommand{\tdtf}{TDT5}

\usepackage{amsmath,amsfonts,bm}

\newcommand{\real}{\mathbb{R}}
\newcommand{\vx}{\bm{x}}
\newcommand{\vz}{\bm{z}}

\newcommand{\CA}{\mathit{CA}}

\newcommand{\dit}{\mathcal{D}}

\usepackage[margin=1in, top=1in]{geometry}
\usepackage{graphicx}
\usepackage{fancyhdr}
\usepackage{hyperref}
\usepackage{xcolor}
\usepackage{titlesec}
\usepackage{tcolorbox}
\usepackage{titling}
\tcbuselibrary{skins}

\usepackage[T1]{fontenc}
\usepackage{tgheros}
\usepackage[utf8]{inputenc}

\usepackage{amsmath,amssymb,bm}
\usepackage{newtxtext}
\usepackage{newtxmath} 

\AtBeginDocument{%

}

\providecommand{\titlefont}{\sffamily\bfseries}
\providecolor{qc_darkblue}{rgb}{0.008,0.063,0.247}
\providecolor{qc_blue}{rgb}{0.164,0.164,0.914}

\titleformat{\title}
{\titlefont\LARGE\color{qc_blue}}{}{0pt}{}

\titleformat{\section}
{\titlefont\Large\bfseries\color{qc_darkblue}}{\thesection}{1em}{}

\titlespacing*{\section}{0em}{1em}{.6em}

\newtcolorbox{titlebox}{
  enhanced,
  colback=white,
  boxrule=0pt,
  opacityback=0,
  opacityframe=0,
  width=0.95\textwidth,
  center
}

\makeatletter
\newcommand{\contactinfo}[1]{\def\@contactinfo{#1}}
\makeatother
\contactinfo{}

\fancypagestyle{titlepage}{
  \fancyhf{}
  \fancyfoot[L]{\footnotesize Qualcomm AI Research is an initiative of Qualcomm Technologies, Inc.}
  \fancyfoot[R]{\footnotesize\thepage}
  
}
\renewcommand\abstract{%
    \setlength{\parskip}{.8em}
    \par
}

\title{Paper Title}
\date{February 23, 2024}
\author{Author1, Author2}
\contactinfo{\{author1, author2\}@qualcomm.com}

\begin{document}
\thispagestyle{titlepage}

\begin{tikz}[remember picture,overlay]
  \node[anchor=north east, inner sep=0pt]
    at ([xshift=5cm,yshift=5.5cm]current page.north east)
    {\includegraphics[width=10cm,angle=-75,origin=c]{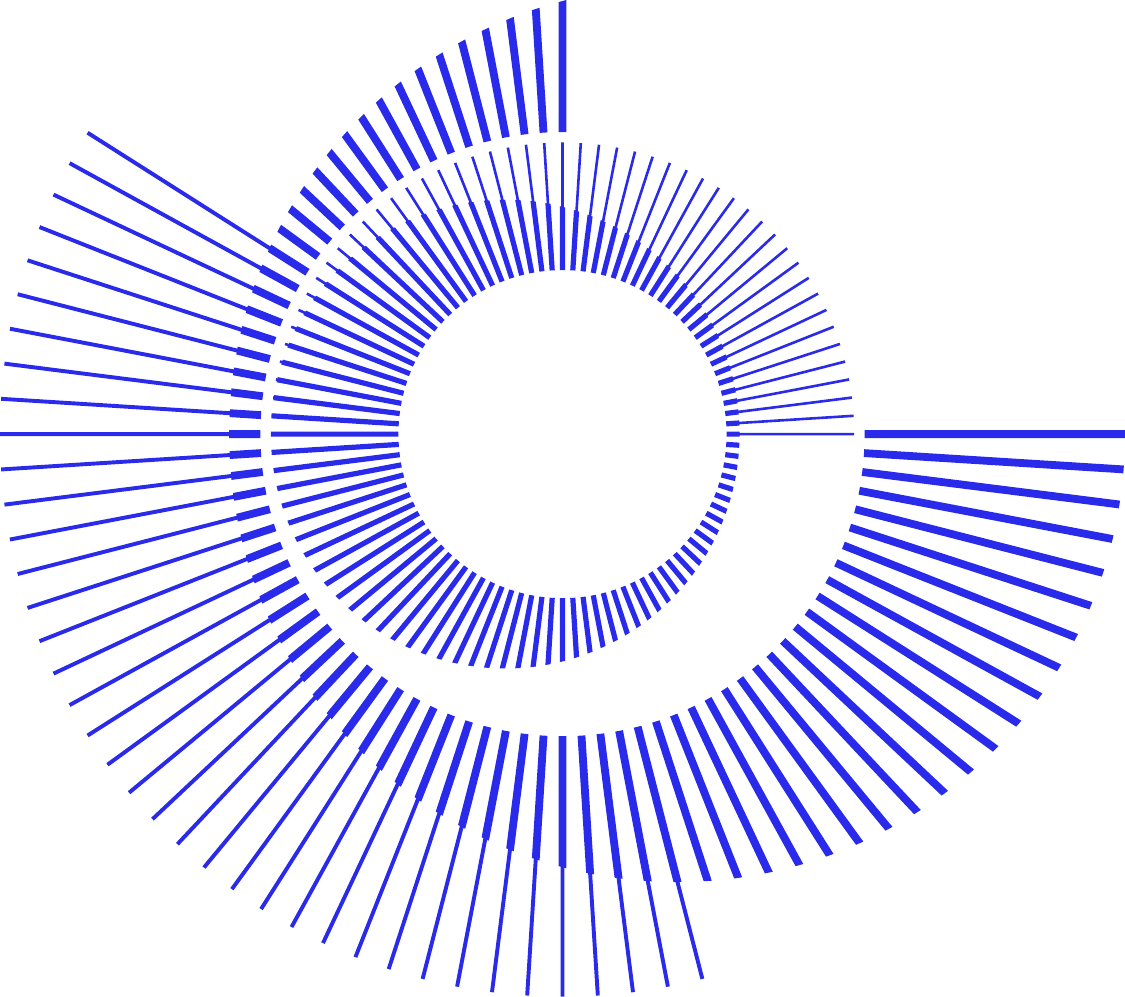}};
\end{tikz}

\begin{figure}[t]
    \vspace*{-1cm}
    \hspace*{-0.6cm} 
    \includegraphics[width=4.0cm]{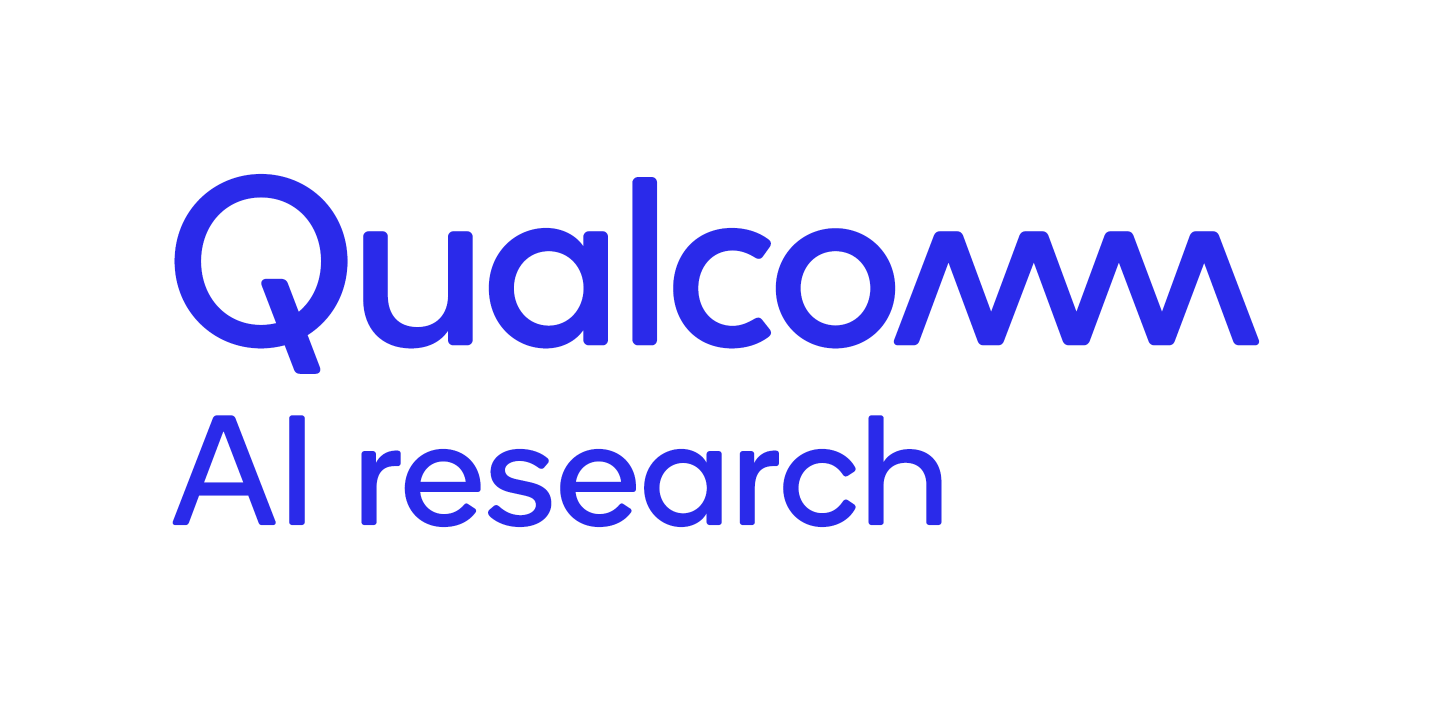}
    \vspace*{-0.5cm}
\end{figure}

\title{\textbf{\name}: \Large{\\Mobile Video Generation using Diffusion Transformer}}
\date{November 6, 2025}
\author{
    Animesh Karnewar,
    Denis Korzhenkov,
    Ioannis Lelekas,
    Adil Karjauv,
    Noor Fathima,
    Hanwen Xiong,
    Vancheeswaran Vaidyanathan,
    Will Zeng,
    Rafael Esteves,
    Tushar Singhal,
    Fatih Porikli,
    Mohsen Ghafoorian,
    Amirhossein Habibian
}
\contactinfo{}

\begin{titlebox}
{\titlefont\huge\bfseries\color{qc_darkblue}\thetitle}\\[1em]

\makeatletter
{\titlefont\color{qc_blue}\@author\par
 \ifx\@contactinfo\@empty
 \else
   \vspace{-.0em}
   {\bfseries\emphfont\@contactinfo}%
 \fi}
\makeatother

\vspace{.5em}

\begin{abstract}
We introduce \textbf{\name}, a text-to-video system capable of generating 2s (49 frames @24 fps) videos at a resolution of 640$\times$1024 directly on a \textbf{\platform} in a record \textbf{$\sim$6.7s} (7 FPS). Differing from existing transformer-based offline text-to-video generation models, \name is the first to have been specifically optimised for mobile hardware to achieve efficient, low-cost, and high-fidelity video synthesis.
We achieve this through four key technical contributions:  
(1) Replacing the original large 4.762B \tfxxl Text-Encoder with a much smaller 0.2B \dtf (DistilT5) with minimal quality loss, enabling the entire model to run without CPU offloading. This is enabled through a novel Text-Encoder Distillation procedure which uses only generative text-prompt data and \textit{does not} require any image or video data.  
(2) Proposing an Asymmetric Decoder Distillation approach which allows us to replace the native codec-latent-VAE decoder with a more efficient one, without disturbing the generative latent-space of the video generation pipeline.  
(3) Pruning of MMDiT blocks within the denoiser backbone based on their relative importance, with recovery of original performance through a two-stage distillation process.  
(4) Reducing the NFE (Neural Functional Evaluation) requirement of the denoiser by performing step distillation using a technique adapted from DMD for \textit{pyramidal} flow-matching, thereby significantly accelerating video generation.
When paired with an optimised SSD1B first-frame image generator and QuickSRNet for $2\times$ super-resolution, our end-to-end \name system becomes a highly parameter (\textbf{4.945B} full model), memory (\textbf{3.5GB} peak RAM usage), and runtime (\textbf{6.7s} E2E latency) efficient mobile-friendly model, while achieving a \textit{VBench} total score of \textbf{81.61}, yielding high-fidelity generated videos.
By enabling low-cost, private, and on-device text-to-video synthesis, \name democratizes AI-based video content creation, empowering creators to generate high-quality videos without reliance on cloud services. Code and model will be made publicly available at our website.
\end{abstract}

\vspace{1em}
\url{https://qualcomm-ai-research.github.io/neodragon} \\

\end{titlebox}

\newcommand{\ignore}[1]{}
\section{Introduction}

Video generation stands at the cusp of becoming the next transformative leap in artificial intelligence. As highlighted in OpenAI’s recent technical report, ``Video Generation Models as World Simulators'' \cite{openai2024videogen}, \ignore{researchers like Bill Peebles argue} it is argued that scaling video generation models could be a promising path toward building general-purpose simulators of the physical world. These models, capable of generating temporally coherent, high-fidelity videos from textual prompts, are not just tools for visual storytelling—they represent a new modality for machines to understand and simulate complex, dynamic environments. This shift could be as foundational as the rise of large language models, unlocking new capabilities in reasoning, planning, and interaction.

Beyond their potential for AGI, text-to-video models are poised to revolutionize creative expression. They offer users the ability to bring ideas to life visually, enabling applications in education, marketing, entertainment, and personal storytelling. The global film industry, valued at over \$136 billion in 2018 when combining box office and home entertainment revenues \cite{wikiFilmIndustry}, exemplifies the scale of opportunity for generative video technologies. Meanwhile, the creator economy—driven by platforms like TikTok, YouTube, and Instagram—has grown into a \$205 billion market as of 2024, with over 165 million new creators joining since 2020 \cite{spiralytics2025creator}. These figures underscore the demand for accessible, high-quality video generation tools that empower both professionals and amateurs alike.

However, the computational demands of state-of-the-art video generation models have so far limited their accessibility. Most systems rely heavily on cloud-based infrastructure, which introduces latency concerns, privacy concerns, and significant operational costs. This reliance creates a barrier for widespread adoption, especially among creators in regions with limited connectivity or creators with limited financial resources. What if we could eliminate this dependency on cloud by enabling on-device video generation? Such a shift would democratize access to foundational generative models, allowing creators everywhere to generate high-quality videos directly on their mobile devices—without needing to upload data to the cloud. This is our main motivation for \name. We take our first stride towards democratising AI-based video content creation and empowering creative minds.

\begin{figure}[t]
  \centering
  \includegraphics[width=\textwidth]{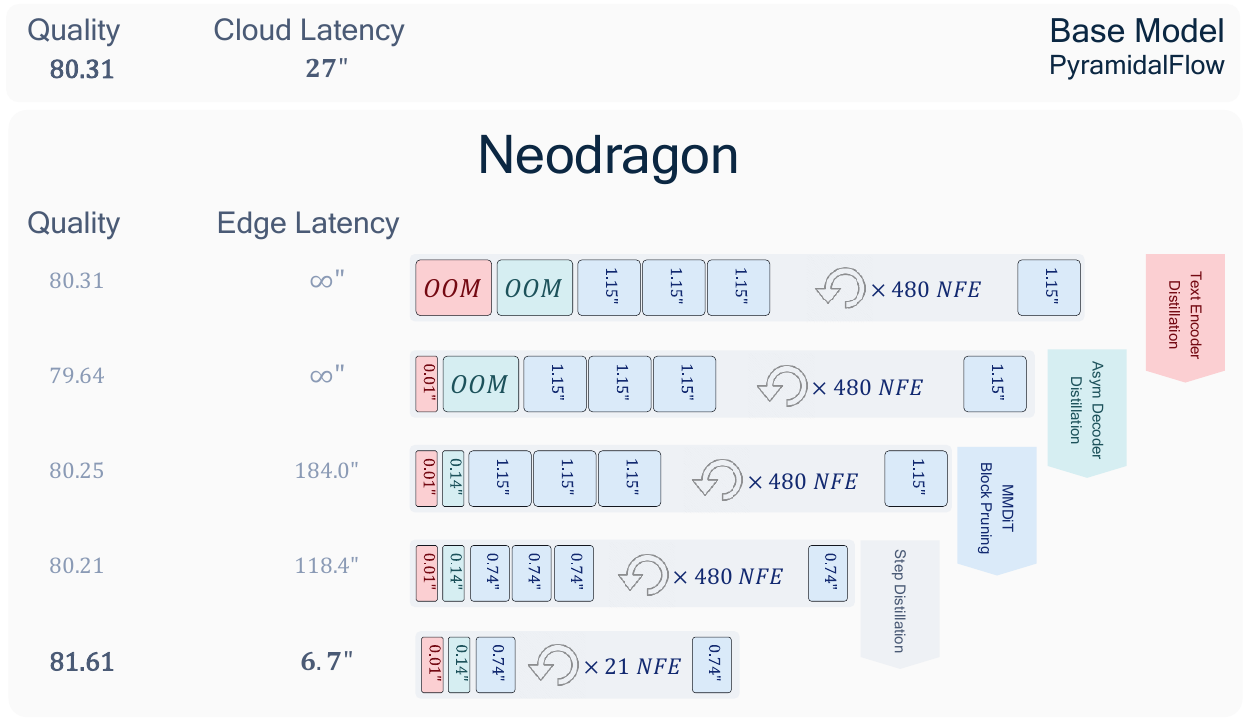}
  \captionof{figure}{An overview of optimisation process steps of \textbf{\name}, our proposed efficient text-to-video generation system designed to run directly on mobile devices powered by \textbf{\platform}.}
  \label{fig:teaser-fig}
\end{figure}

Recent advances in video diffusion modeling have seen a shift from traditional U-Net architectures \cite{bar-tal2024lumiere, ruan2023mmdiffusion, ho2022videoddpm} to Transformer-based designs \cite{zhang2025tora, peebles2022dit, HaCohen2024LTXVideo, cheng2024videodit}, with Diffusion Transformers (DiTs) \cite{peebles2022dit} emerging as the new state-of-the-art due to their superior scalability and performance in generating temporally coherent and high-fidelity video content~\cite{melnik2024survey, thor2024unetvsdit}. On the other hand, most recent works on optimizing the video diffusion models for on-device executions~\cite{wu2025snapgen,benyahia2024mobilevd, karjauv2024movie, kahatapitiya2024object} focus exclusively on spatio-temporal U-Nets, leaving a notable gap in the literature regarding DiT-based models and methods to tailor them for on-device execution. To the best of our knowledge, Mobile Video DiT by Wu et al.~\cite{wu2025taming}, and On-Device Sora by Kim et al.~\cite{kim2025device}, are the only recent and parallel works that specifically attempt to address the challenges of deploying Video DiTs on mobile devices. 

\begin{figure}[!t]
  \centering
  \includegraphics[width=\textwidth]{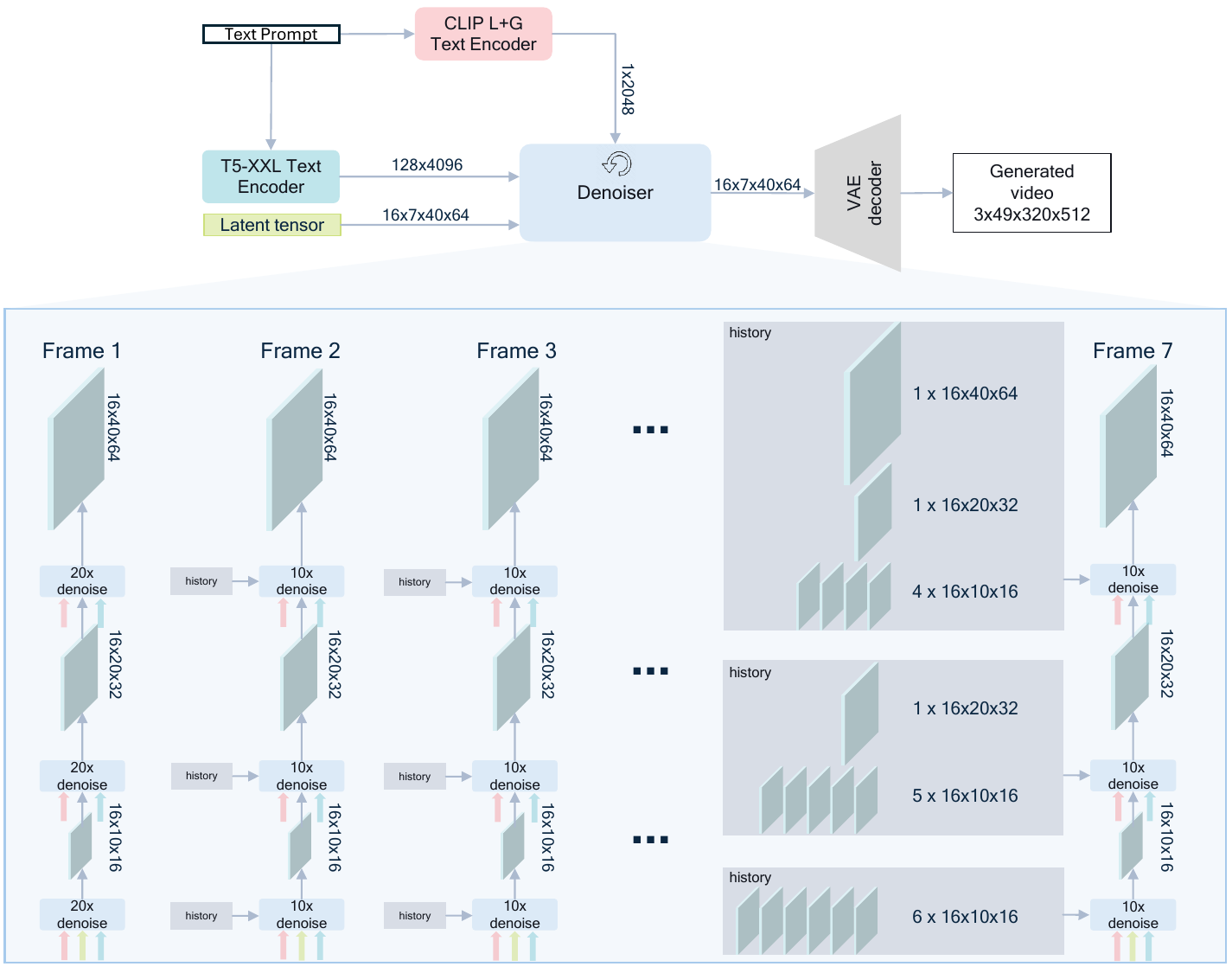}
  \captionof{figure}{\textbf{Overview of the Pyramidal Autoregressive Video Diffusion Pipeline}. The pyramidal autoregressive video diffusion scheme \cite{jin2024pyramidal} differs from the conventional latent-diffusion in how the the latent-video frames are generated (iteratively denoised). The latent frames are autoregressively generated one-by-one by denoising the curent frame while conditioning on the past history. A spatio-temporal pyramid is applied in the denoising process as: firstly the denoising of the current frame starts from a lower resolution and proceeds to reach the highest native latent-resolution; and secondly, each denoising step is conditioned on past history, where the frames from the further past are spatially downsampled.}
  \label{fig:base_pipeline_illustration}
\end{figure}

In this work, we present a novel DiT-based text-to-video generation system optimised for mobile hardware. Our system is designed to run efficiently on modern smartphone platforms as well as ARM laptops, achieving low latency  while maintaining competitive video quality (see fig.~\ref{fig:teaser-fig}). Our main contributions are summarised below:

\begin{enumerate}
    \item We introduce \textbf{\name}, our end-to-end text-to-video generative system, integrating multiple video DiT components we optimised to run directly on the mobile devices powered by \platform, generating high-quality videos efficiently. The final optimised system compares competitively to various offline cloud-based diffusion models (ref sec.~\ref{sec:e2e_integration}). 
    
    \item We propose a \textbf{Text-Encoder Distillation} framework that compresses the 4.762B-parameter \tfxxl model by 35× into a lightweight encoder 0.130B-parameter \dtf (DistilT5) using a newly trained 0.130B-parameter $\CA$ (ContextAdapter) module; while achieving this reduction with no significant drop in overall video generation quality. Our proposed training process does not require any image or video supervision (ref subsec.~\ref{subsec:text_enc_distil}).

    \item We introduce an \textbf{Asymmetric Decoder Distillation} strategy that replaces 
    the native codec-latent-VAE decoder with a new mobile-friendly decoder while preserving the original generative latent-space, achieving over 20× parameter reduction and actual on-device execution with negligible impact on video generation quality (ref subsec.~\ref{subsec:asymm_dec_distil}).

    \item We propose a novel \textbf{MMDiT Block Pruning} strategy for the denoiser backbone that reduces parameters from 2.009B to 1.499B (\textgreater25\% reduction) with minimal quality loss, using a three-step process of block importance scoring, followed by data based fine-tuning, and finally Full Teacher model distillation (ref subsec.~\ref{subsec:block_pruning}).

    \item We are the first to adapt DMD (Distribution Matching Distillation) based \textbf{Step Distillation} method for the \textit{Pyramidal} Flow-Matching \cite{jin2024pyramidal} on video diffusion, managing to reduce the number of NFEs from 480 to 21 (\textgreater95\% reduction) without affecting the VBench score (ref subsec.~\ref{subsec:step_distil}).
\end{enumerate}

\section{Mobile Efficiency Requirements}

Table \ref{tab:t2v_model_compare} summarises recently released text-to-video diffusion models that are publicly available. For building a mobile text-to-video system, we consider the constraints imposed by mobile hardware platforms. The primary limitations are twofold: (\textit{i}) the total model size, which must fit within the DRAM capacity of the target device, and (\textit{ii}) the computational complexity, as this directly influences both the power consumption and the inference latency. Regarding the first constraint, as shown in Table \ref{tab:t2v_model_compare}, all models except Hunyuan Video \cite{hunyuan2025} offer relatively compact checkpoints, providing some flexibility in model selection. However, the second constraint—compute complexity—requires a more careful analysis. In the following discussion, we explain how the two features of the model from \baseline, namely Causal Attention and Token-Savings (in the form of Pyramid) contribute to the mobile-friendliness of it, and thereby motivating why we select it as the foundation of our \name~system.

\begin{table}[!t]
\centering
\caption{\textbf{Comparison of Recent Text-to-Video Models}. Rows are sorted by release date (ascending).}
\label{tab:t2v_model_compare}
\resizebox{\linewidth}{!}{%
\begin{tabular}{|l|l|l|c|c|c|c|}
\hline
\textbf{Model} & \textbf{Release} & \textbf{Generation Resolution} & \textbf{Model-Size} & \textbf{VBench Total} & \textbf{Causal?} & \textbf{Token-Savings?} \\
\hline
Cogvideo-X~\cite{yang2024cogvideox}& 19 Sep 2024 & \texttt{49$\times$768$\times$1360}& 2.0B & 81.55 & \redcross  & \redcross  \\
\hline
\cellcolor{tabbaseline}\baseline~\cite{jin2024pyramidal}&\cellcolor{tabbaseline}29 Oct 2024 & \cellcolor{tabbaseline}\texttt{49$\times$384$\times$512} &\cellcolor{tabbaseline}2.0B &\cellcolor{tabbaseline}81.56 &\cellcolor{tabbaseline}\greentick &\cellcolor{tabbaseline}\greentick \\
\hline
Hunyuan Video~\cite{hunyuan2025}   & 3 Dec 2024  & \texttt{129$\times$544$\times$960}& 13.0B  & 85.09 & \redcross  & \redcross  \\
\hline
LTXVideo~\cite{HaCohen2024LTXVideo}& 30 Dec 2024 & \texttt{81$\times$512$\times$768} & 2.0B   & 82.30 & \redcross  & \redcross  \\
\hline
Wan~\cite{wan2025}                 & 25 Feb 2025 & \texttt{81$\times$480$\times$832} & 1.3B & 84.26 & \redcross  & \redcross  \\
\hline
OpenSORA~\cite{lin2024opensora}    & 13 Mar 2025 & \texttt{129$\times$768$\times$1365}& 1.1B & 77.70 & \redcross  & \greentick \\
\hline
\end{tabular}
}
\end{table}

Latent video diffusion models \cite{blattmann2023stable, yang2024cogvideox, jin2024pyramidal,hunyuan2025,HaCohen2024LTXVideo,wan2025,lin2024opensora}, building on the framework introduced by Rombach et al.~\cite{rombach2022high} for images, generate videos in two stages. First, a codec-latent-VAE compresses the input from pixel space to a lower-dimensional latent space using architectures such as VQ-VAE~\cite{razavi2019generating} or VQ-GAN~\cite{yu2021vector}. Second, a diffusion model is trained on these compressed latents, enabling efficient yet high-quality generation. This design balances computational efficiency with generative capacity for high-resolution video synthesis. Intuitively, the diffusion model handles the more challenging task of composing scene layout, objects, and their spatial relationships, while the VAE focuses on reconstructing textures and perceptual details.
Formally, during training, an input video $\vx \in \real^{T \times H \times W \times 3}$ is mapped to a latent representation $\vz \in \real^{t \times h \times w \times c}$ via a spatio-temporal encoder: $\vz = \mathcal{E}_{\text{enc}}(\vx)$, where $t = T / f_t$ is the number of latent frames, and $h = H / f_s$, $w = W / f_s$ are the spatial dimensions reduced by a factor $f_s$ (typically $8$), and $c$ are the number of latent channels. The temporal compression factor $f_t$ is usually $4$ or $8$, slightly different from $f_s$ (Although for \baseline~it is also 8). During inference, the diffusion model generates a latent video $\hat{\vz} \in \real^{t \times h \times w \times c}$ starting from Gaussian noise, which is then decoded into RGB frames by: $\hat{\vx} = \mathcal{E}_{\text{dec}}(\hat{\vz})$.

As summarised in Table \ref{tab:t2v_model_compare}, most open-sourced video diffusion models—except for \baseline—employ fully bidirectional self-attention \cite{vaswani2017attention} within transformers to generate or denoise the $t\!\times\!h\!\times\!w\!\times\!c$ latent representations. Due to the all-to-all nature of self-attention, the computational complexity of such bidirectional video transformers is given by $\mathcal{C}_{\text{bi}}$ in Equation~\ref{eq:bidirectional}. Here we can amortize the computational complexity as the number of dot-products computed by a single self-attention~\cite{vaswani2017attention} operation in the Transformer network. As we know, the number of dot-products are equal to the square of the number of tokens input to the attention operation, due to the all-to-all nature of the operation.  In contrast, a causal frame-by-frame transformer operating on the same latent size has complexity $\mathcal{C}_{\text{causal}}$, as derived in Equation~\ref{eq:causal}. Since causal transformers achieve approximately a $2\times$ reduction in compute compared to fully bidirectional counterparts (see eq.~\ref{eq:bd_caus_comp}), \baseline~\cite{jin2024pyramidal} emerges as a desirable foundation model for our system. Apart from the $2\!\times\!$ speedup, the ability to generate videos in a streaming fashion on the fly is another characteristic of the causal attention transformer which adds to the desirability. Now going one step further, we show that the temporal pyramidal conditioning for the causal frame-by-frame generation of \baseline, actually gives us 32$\times$ compute saving as opposed to the 2$\times$ saving of vanilla non-pyramidal causal model (see fig.~\ref{fig:base_pipeline_illustration}). 

\begin{align}
\label{eq:bidirectional}
\textbf{Bidirectional Attention:} \quad
\mathcal{C}_{\text{bi}} &= (hwt)^2 \\[10pt]
\textbf{Causal Attention:} \quad
\mathcal{C}_{\text{causal}} &= \sum_{k=1}^{t} \underbrace{(h \cdot w)}_{\text{tokens in frame $k$}} \times \underbrace{(h \cdot w \cdot k)}_{\text{tokens in frames $1..k$}} \\[6pt]
&= \sum_{k=1}^{t} (hw)^2 \cdot k \\[6pt]
&= (hw)^2 \sum_{k=1}^{t} k \\[6pt]
&= (hw)^2 \cdot \frac{t(t+1)}{2} \label{eq:causal} \\[10pt]
\label{eq:bd_caus_comp}
\quad
\text{\textbf{Speedup}}_\text{\textbf{causal}}
\;=\;
\frac{\mathcal{C}_{\text{bi}}}{\mathcal{C}_{\text{causal}}} &= \frac{(hw)^2 t^2}{(hw)^2 \cdot \frac{t(t+1)}{2}} = \frac{2t}{t+1} \approx 2\!\times\! \text{ as $t \to \infty$}
\end{align}

\paragraph{\textit{Temporally Pyramidal Causal} latent generation (general $S$; $t>S$).}
We use a temporal pyramid with $S$ stages indexed by $i\in\{0,\ldots,S\!-\!1\}$.
Stage $i$ corresponds to a spatial resolution that is downsampled by $2^i$ per dimension
relative to the highest resolution. If a full-resolution frame has $M = h\cdot w$ tokens, then the number of tokens contributed by a frame at stage $i$ is
\[
M_i \;=\; \frac{M}{4^{\,i}}, \qquad i=0,1,\ldots,S\!-\!1,
\]
since each $2\times$ reduction per spatial dimension reduces the token count by a factor of $4$.
We refer to stage $0$ as the highest (full-resolution) stage and stage $S\!-\!1$ as the lowest stage. For a query at frame $k$ and a history frame $j\le k$, let the temporal distance be $d:=k-j$.
The number of tokens contributed by this particular history frame are:
\[
T(d) \;=\;
\begin{cases}
M, & d=0 \quad(\text{self}),\\[2pt]
\dfrac{M}{4^{\,d-1}}, & 1 \le d \le S-1,\\[8pt]
\dfrac{M}{4^{\,S-1}}, & d \ge S.
\end{cases}
\]
Each query frame has $M$ query tokens, so the dot-product cost contributed by a $(k,j)$ pair is $M \cdot T(d)$.
Summing over all ordered pairs $(k,j)$ with $1\le j\le k\le t$ is equivalent to summing over distances $d$
and counting how many pairs have that distance:
for a fixed $d$, there are exactly $(t-d)$ pairs $(k,j)$ with $k-j=d$.

\paragraph{Total complexity (general $S$).}
Let $r=\tfrac{1}{4}$ be the token downsampling factor, and define the finite sums
\[
A(S):=\sum_{m=0}^{S-2} r^{m}
=\frac{1-r^{S-1}}{1-r}
=\frac{4}{3}\Big(1-4^{-(S-1)}\Big),
\qquad
D(S):=\sum_{d=1}^{S-1} d\,r^{\,d-1}
=\frac{1-(S)r^{S-1}+(S-1)r^{S}}{(1-r)^2}.
\]
With $u:=t-S$ (and $t>S$ so $u\ge 1$), we obtain
\begin{align}
\label{eq:Cpyr_general}
\mathcal{C}_{\text{pyr}}(t,S)
&= \sum_{k=1}^{t}\sum_{j=1}^{k} M\cdot T(k-j)
= M^2\!\left[
\underbrace{t}_{\text{self (}d=0\text{)}}\;
+\;
\underbrace{\sum_{d=1}^{S-1} (t-d)\,r^{\,d-1}}_{\text{geometric ramp }(d=1..S-1)}
\;+\;
\underbrace{r^{\,S-1}\sum_{d=S}^{t-1}(t-d)}_{\text{bulk at lowest stage }(d\ge S)}
\right] \nonumber\\[4pt]
&= M^2\!\left[
t \;+\; t\,A(S) \;-\; D(S) \;+\; \frac{u(u+1)}{2\cdot 4^{\,S-1}}
\right].
\end{align}
\emph{Asymptotically} as $t\to\infty$ (with fixed $S$),
\begin{equation}
\label{eq:Cpyr_asym}
\mathcal{C}_{\text{pyr}}(t,S)
= \frac{M^2}{2\cdot 4^{\,S-1}}\,t^2 \;+\; \mathcal{O}(t),
\qquad\Longrightarrow\qquad
\text{Speedup}_\text{pyr}(S) = \frac{\mathcal{C}_{\text{bi}}}{\mathcal{C}_{\text{pyr}}(t,S)}
~\xrightarrow[t\to\infty]{}~
2\cdot 4^{\,S-1}.
\end{equation}

\paragraph{Specialisation to $S=3$ (matches Fig.~\ref{fig:base_pipeline_illustration}).}
For $S=3$, equation \ref{eq:Cpyr_asym} becomes: 
\begin{align}
    \text{\textbf{Speedup}}_\text{\textbf{temporal}} = \text{Speedup}_\text{pyr}(S=3) = 2\cdot4^{(3 - 1)} = 32\!\times\!.
\end{align}

\noindent The $32\!\times$ compute saving from the \textit{Temporally Pyramidal Causal} attention is already a major boost, but the \baseline~model goes further by also denoising each frame in a \textit{spatial pyramid} (coarse-to-fine) fashion. This spatial pyramidal structure is orthogonal to the temporal pyramid and provides an additional speedup. We now derive this spatial speedup and then combine it with the temporal speedup to obtain the total compute savings over full bidirectional attention (see Fig.~\ref{fig:base_pipeline_illustration}).

\paragraph{\textit{Spatial pyramid setup.}}
Assume that the denoising process allocates fractions $p_i$ of the total denoising steps to each stage, with $\sum_{i=0}^{S-1} p_i = 1$.

\paragraph{Per-frame cost scaling.}
At stage $i$, both the query and the effective K/V token counts scale by $1/4^i$ relative to full resolution. Since attention cost is bilinear in queries and keys, the per-frame cost at stage $i$ scales as
\[
\text{Cost factor at stage $i$} \;\propto\; \frac{1}{4^i}\cdot\frac{1}{4^i} = \frac{1}{16^i}.
\]
Thus, if $\mathcal{C}_{\text{temp}}^{(k)}$ denotes the per-frame cost under the \textit{Temporally Pyramidal Causal} setup (with queries at full resolution), then the spatially adjusted per-frame cost is
\[
\mathcal{C}_{\text{spatial-temp}}^{(k)}
= \Bigg(\sum_{i=0}^{S-1} \frac{p_i}{16^{\,i}}\Bigg)\, \mathcal{C}_{\text{temp}}^{(k)}
= \beta_S(\mathbf{p})\, \mathcal{C}_{\text{temp}}^{(k)},
\qquad
\beta_S(\mathbf{p}) := \sum_{i=0}^{S-1} \frac{p_i}{16^{\,i}}.
\]

\paragraph{Spatial speedup.}
The relative compute multiplier in the spatial dimension is $\beta_S(\mathbf{p})<1$, so the speedup is
\[
\boxed{\;\text{Speedup}_{\text{spatial}}(\mathbf{p}) = \frac{1}{\beta_S(\mathbf{p})}\;}.
\]

\paragraph{Uniform allocation across stages.}
If denoising steps are split uniformly across stages, $p_i=\tfrac{1}{S}$, then
\[
\beta_S
= \frac{1}{S}\sum_{i=0}^{S-1}\frac{1}{16^{\,i}}
= \frac{1}{S}\cdot\frac{1-16^{-S}}{1-\frac{1}{16}}
= \frac{16}{15S}\Big(1-16^{-S}\Big),
\qquad
\text{Speedup}_{\text{spatial}}
= \frac{15S}{16(1-16^{-S})}.
\]
For large $S$, this approaches $\beta_S\approx \tfrac{16}{15S}$ and $\text{Speedup}_{\text{spatial}}\approx \tfrac{15}{16}S$.

\paragraph{Specialisation to $S=3$.}
With three spatial stages and uniform allocation $p_i=\tfrac{1}{3}$,
\[
\beta_3
= \frac{1}{3}\Big(1+\frac{1}{16}+\frac{1}{256}\Big)
= \frac{273}{768}
\approx 0.3555,
\qquad
\text{\textbf{Speedup}}_{\text{\textbf{spatial}}}
= \frac{1}{\beta_3}
= \frac{768}{273}
\approx 2.81\!\times\!.
\]

\paragraph{Combined spatio-temporal speedup.}
The temporal pyramid (with $S=3$ stages) yields an asymptotic speedup of
\[
\text{\textbf{Speedup}}_{\text{\textbf{temporal}}} \approx 32\!\times\!
\]
relative to full bidirectional attention.
The spatial pyramid (with $S=3$ stages and uniform allocation) yields
\[
\text{\textbf{Speedup}}_{\text{\textbf{spatial}}} \approx 2.81\!\times\!
\]
relative to the temporal-only baseline.
Since these optimisations act on orthogonal dimensions (temporal vs spatial),
the combined speedup is multiplicative:
\[
\boxed{
\text{\textbf{Speedup}}_{\text{\textbf{combined}}} \approx 32 \times 2.81 \approx 90\!\times\!.
}
\]


Thus, a \emph{Spatio-temporally Pyramidal Causal} latent generation setup can reduce
the dominant attention complexity by nearly two orders of magnitude ($90\!\times\!$) compared
to a full-resolution, fully bidirectional attention. These efficiency gains are not merely theoretical; they enable practical scaling of autoregressive video diffusion to longer sequences and higher resolutions without prohibitive compute costs. For this reason, we adopt \textbf{\baseline} \cite{jin2024pyramidal} as the foundation for our \textbf{\name} system, leveraging its hierarchical structure to deliver both computational efficiency and strong generative performance.

For the pre-super-res output, the spatial resolution targeted in our setup is \texttt{[320$\times$512]}. Thus the starting point for us is the \baseline model, which achieves a total VBench score of \textbf{80.31}. This score is obtained using the inference parameters provided by the authors in their released code, applied to the 480p low-resolution checkpoint—note that no official VBench score is reported for this checkpoint. The officially reported score for \baseline corresponds to its higher-resolution 720p checkpoint, which achieves \textbf{81.72}. 
It is important to acknowledge that the \baseline~model, which we adopt as our foundation, was trained with comparatively fewer GPU hours than some of the more resource-intensive open-source models such as Wan \cite{wan2025}, CogVideoX \cite{yang2024cogvideox}, and LTX-Video \cite{HaCohen2024LTXVideo}. Consequently, any inherent limitations in Pyramid-Flow’s generations are inherited by our system. However, the suite of optimisations we introduce to enable efficient deployment on our platform is broadly applicable and can be extended to other models as well.

\subsection{Experimental Protocol}

Our experimental setup is designed to evaluate the proposed optimisations for mobile hardware deployment in a controlled and reproducible manner. We start from the original \baseline model \cite{jin2024pyramidal}, which serves as the backbone for our system due to its hierarchical spatio-temporal design. On top of this, we progressively integrate the components detailed in the following section (ref sec. \ref{sec:method}).

\paragraph{Datasets.}
To train our \textbf{Text-Encoder Distillation} framework (subsec.~\ref{subsec:text_enc_distil}), we curated a diverse corpus of approximately 1.4M generative text prompts. These were sourced from CommonText \cite{wang2025scaling}, DiffusionDB \cite{wangDiffusionDBLargescalePrompt2022}, a high-aesthetic subset of LAION (score $>$ 6.5) \cite{schuhmann2022laionb}, and T2ICompbench \cite{huang2023t2icompbench}.  
For \textbf{Asymmetric Decoder Distillation} (subsec.~\ref{subsec:asymm_dec_distil}) and the two-stage fine-tuning of our \textbf{MMDiT Block-Pruning} approach (subsec.~\ref{subsec:block_pruning}), we use $\sim$253K videos with captions from the Stock Videos subset (Mixkit, Pexels, Pixabay) of OpenSora \cite{lin2024opensora}, combined with $\sim$87K video-text pairs from Panda-70M \cite{chen2024panda}, totaling $\sim$350K samples.  
Finally, for \textbf{Step Distillation} (subsec.~\ref{subsec:step_distil}), we generate synthetic videos for the curated $\sim$350K text prompts using the 480p checkpoint of our base Pyramidal-Flow model \cite{jin2024pyramidal}.

\paragraph{Metrics.}
We evaluate all experiments using the VBench suite \cite{huang2024vbench}, which provides a comprehensive set of metrics for video generation quality and consistency. \tabcapsfx

\section{\textbf{\name}}
\label{sec:method}
\subsection{Text-Encoder Distillation}
\label{subsec:text_enc_distil}

\begin{figure}[!t]
  \centering
  \begin{overpic}[width=\textwidth]  {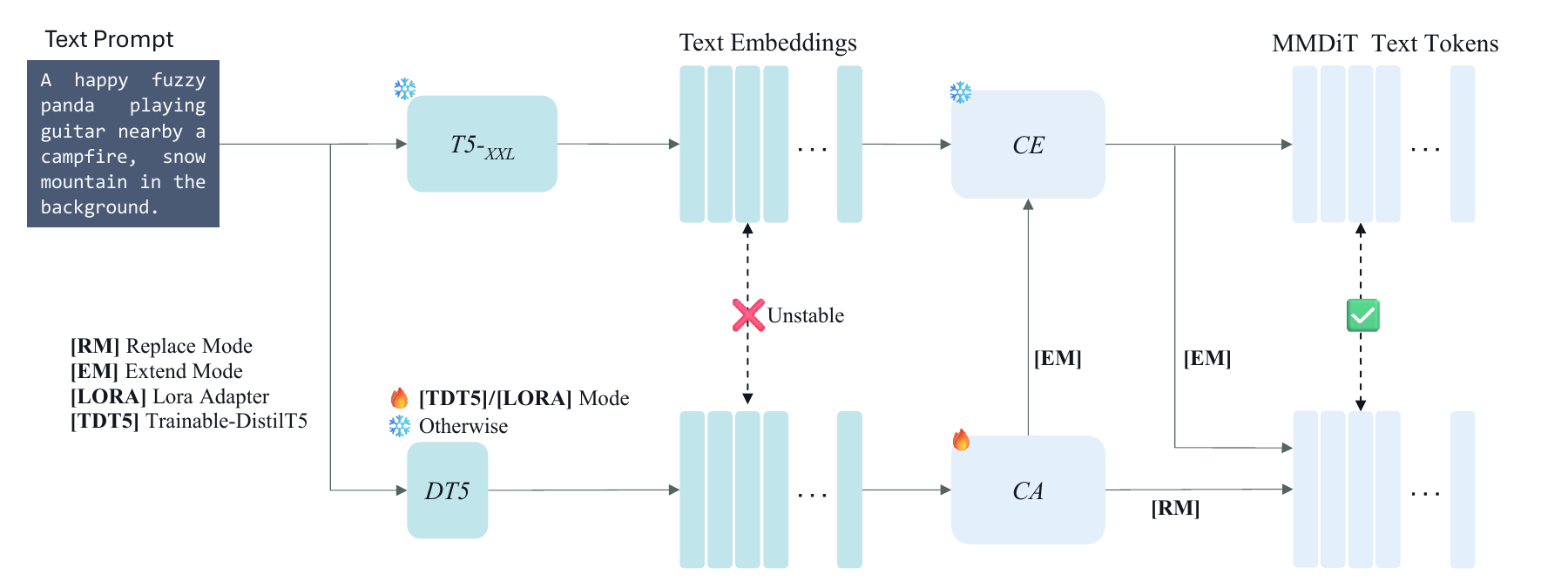}
  \put(88.5,17){\parbox[c]{2cm}{\footnotesize Stable \\ $\mathcal{L}_\text{distil}(t, \hat{t})$}}
  \end{overpic}
  \captionof{figure}{\textbf{Overview of the proposed Text-Encoder Distillation framework}. The original large-scale text-encoder \tfxxl is distilled into a light-weight model via a trainable $\mathit{CA}$ (ContextAdapter) module, using a combination of MSE and Cosine Distance loss to align the embeddings. Multiple modes are supported in our framework -- Replace Mode \textbf{[RM]}: where the new $\mathit{CA}$ \textit{replaces} the original $\mathit{CE}$ (ContextEmbedder); Extend Mode \textbf{[EM]}: where the new $\mathit{CA}$ \textit{extends} the original $\mathit{CE}$; Lora Mode \textbf{[LORA]}: Where the $\mathit{CA}$ is not a separate MLP, but LoRA \cite{hu2022lora} layers on top of the \dtf text-encoder; and, we allow training the smaller text-encoder v/s keeping it frozen via \textbf{[\tdtf]} (\textbf{Trainable-\dtf}) mode.
  }
  \label{fig:text-encoder-illustration}
\end{figure}

Having established the efficiency constraints of our mobile text-to-video generation system, we begin by identifying the baseline hardware latency value for optimisation. Our initial plan was to port the \baseline~model \cite{jin2024pyramidal} onto the SoC platforms powered by \platform~(such as mobile SoC \mobilesoc~or the laptop SoC~\laptopsoc), irrespective of however long the model took for generating a \texttt{[49 x 320 x 512]} video. However, this approach encountered the critical challenge of the large size of the native Text-Encoder \tfxxl. The \tfxxl, with a parameter count of 4.726 billion, overshoots the total model footprint requiring CPU offloading for on-device execution. Because of this, we were unable to obtain reliable latency profiling through our simulation tools. We therefore boiled down our aim to a single guiding question: \emph{is the full capacity of \tfxxl actually necessary for high-quality text-to-video generation?} A direct operational corollary is whether a much smaller encoder can be substituted without perceptible fidelity loss. Motivated by distillation results in text-to-image and vision–language systems~\cite{wang2025scaling,ma2025x2i,zhang2025vlv,zhao2023mobilediffusion}—which suggest large encoders are under-utilised for short, descriptive prompts—we hypothesise that text-to-\emph{video} models impose similarly shallow semantic demands. Building on \dtf (DistilT5)~\cite{wang2025scaling}, we propose a prompt-only Text-Encoder distillation framework tailored to video generation.

Figure~\ref{fig:text-encoder-illustration} illustrates our proposed distillation framework. A direct attempt to train the \dtf model to replicate the text embeddings from the larger \tfxxl model leads to an unstable optimisation. This isn’t surprising because compressing the full spectrum of text-understanding capabilities from a large model into a smaller one is an unrealistic endeavour. Fortunately, our goal is more focused: we only need to distill the aspects of text understanding that are relevant for video generation. To achieve this, we incorporate the $\mathit{CE}$ (ContextEmbedder) from the model, which operates within the MMDiT's namespace. The $\mathit{CE}$ is responsible for transforming the extracted text embeddings from \tfxxl into tokens for conditioning the MMDiT denoiser. And, in order to learn the video generation specific adaption, we introduce a new learnable module called $\mathit{CA}$ (ContextAdapter) into the pipeline. Our training objective combines MSE (Mean Squared Error) and Cosine Distance losses between the predicted conditioning tokens and the ground truth MMDiT text tokens ensuring the distilled model learns the most relevant semantic cues for video generation.
\begin{align}
    &\mathcal{L}_\text{distil}(t, \hat{t}) := w_\text{mse}\left\| t - \hat{t} \right\|_2^2 + w_\text{cd}\left(1 - \frac{t.\hat{t}}{|t|.|\hat{t}|}\right) \label{eq:text_encoder_loss}\\
    &\text{where, } t = \mathit{CE}(T5_\text{XXL}(\texttt{prompt})) \text{ and, } \hat{t} = \mathit{CA}(DT5(\texttt{prompt})) \label{eq:text_encoder_def}
\end{align}

The distillation framework supports multiple configurations, each tailored to explore a specific path over the experimental design space. The ground truth $\mathit{CE}$ (ContextEmbedder) is a single linear layer, serving as a fixed reference throughout. In contrast, the newly introduced $\mathit{CA}$ (ContextAdapter) is a more expressive 4-layer MLP with skip connections at every layer, designed to learn the task-specific adaptations. The framework operates in four different modes: \textbf{[RM]} Replace-Mode, where $\mathit{CA}$ replaces $\mathit{CE}$ entirely; \textbf{[EM]} Extend-Mode, where $\mathit{CA}$ complements $\mathit{CE}$ and both of them are cascaded during inference; \textbf{[\tdtf]}, which makes the \dtf model trainable within the pipeline; and \textbf{[LORA]}, which replaces the MLP-based $\mathit{CA}$ with LoRA \cite{hu2022lora} adapter layers on top of \dtf. Throughout all modes, \tfxxl and $CE$ remain frozen to provide consistent ground truth signals for distillation. The $\mathit{CA}$ is always trainable, while \dtf is only updated in \textbf{[\tdtf]} mode, and in \textbf{[LORA]} mode, only the adapter layers are trainable.

The setup was trained using the Adam optimiser with a learning rate of 3e-3, decayed via a cosine schedule to 3e-5. Training was conducted over 24,000 iterations on four 80GB H100 GPUs, with a batch size of 512 per GPU, resulting in a total batch size of 2048. The complete training process took approximately 16 hours on an average for all the four modes. As our default value, we set $w_\text{mse}=1.0$ and $w_\text{cd}=0.1$, which we found to work best empirically, and provide the ablation over these for the \textbf{[RM]} in Figure~\ref{fig:tex_enc_abl_lossweight}. This setup enabled efficient convergence of the distilled encoder using only text data, without requiring any image or video supervision.

\begin{figure}[!t]
  \centering
  \includegraphics[width=0.9\textwidth]{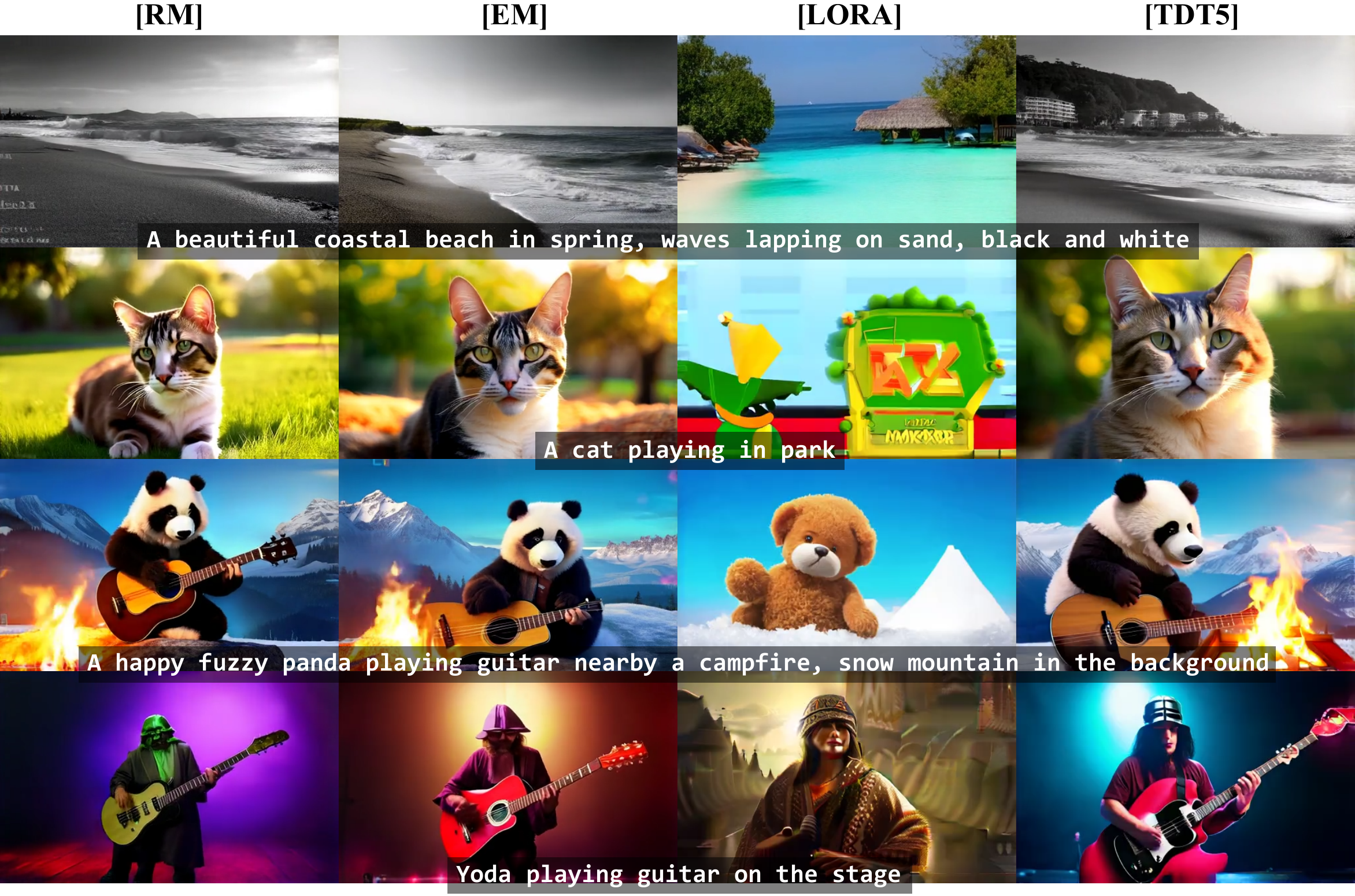}
  \captionof{figure}{\textbf{Qualitative Evaluation of Text-Encoder Distillation.} We visualise randomly selected frames from the generated \texttt{[49×320×512]} videos corresponding to the adjacent text prompts, across the four modes supported by our Text-Encoder Distillation framework: \textbf{[RM]}, \textbf{[EM]}, \textbf{LORA}, and \textbf{[\tdtf~]}.}
  \label{fig:text-encoder-qualitative}
\end{figure}

\begin{figure}[!t]
  \centering
  
  \begin{subfigure}[t]{0.48\textwidth}
    \centering
    \includegraphics[width=\textwidth]{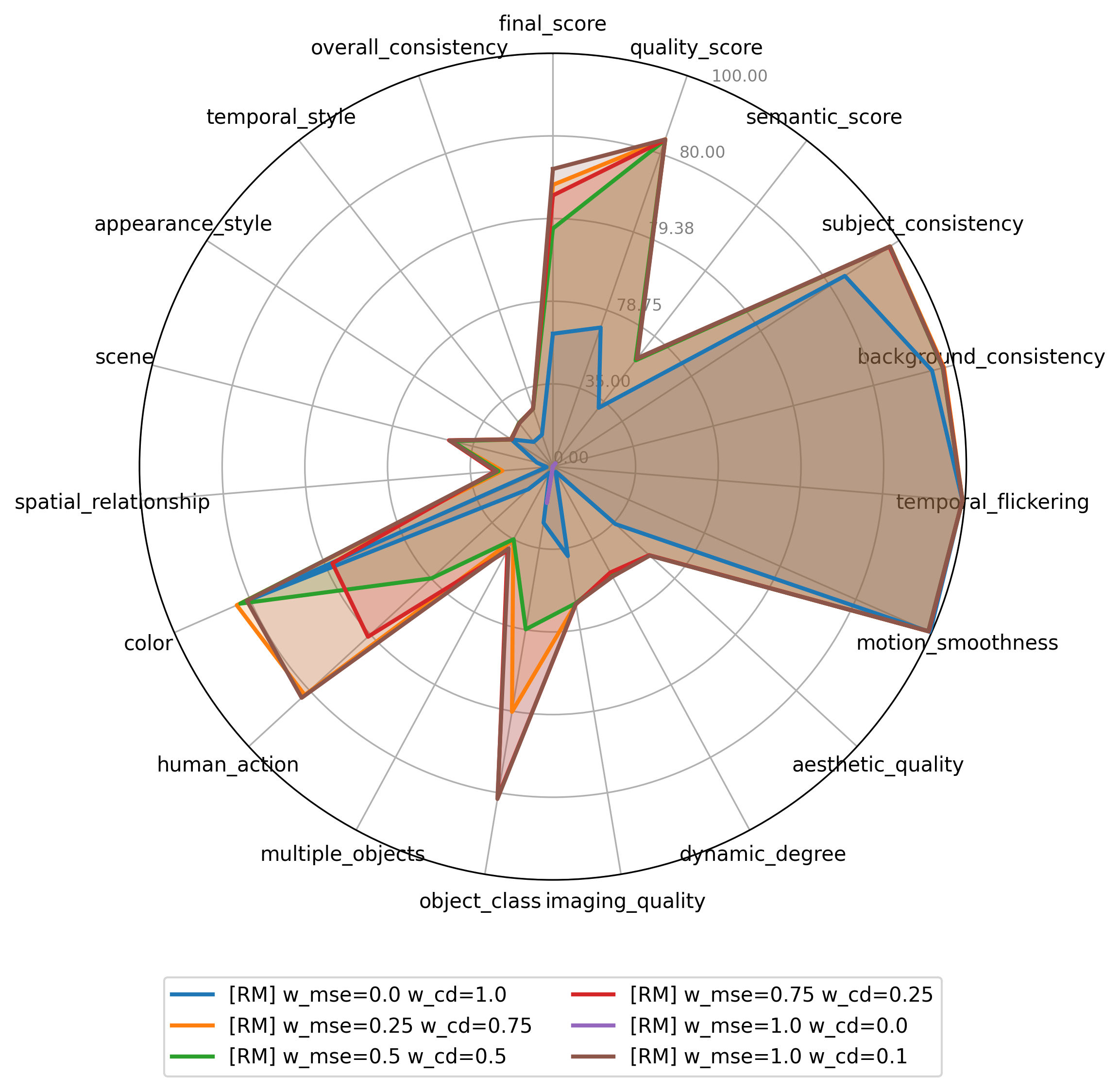}
    \caption{}
    \label{fig:tex_enc_abl_lossweight}
  \end{subfigure}
  \begin{subfigure}[t]{0.48\textwidth}
    \centering
    \includegraphics[width=\textwidth]{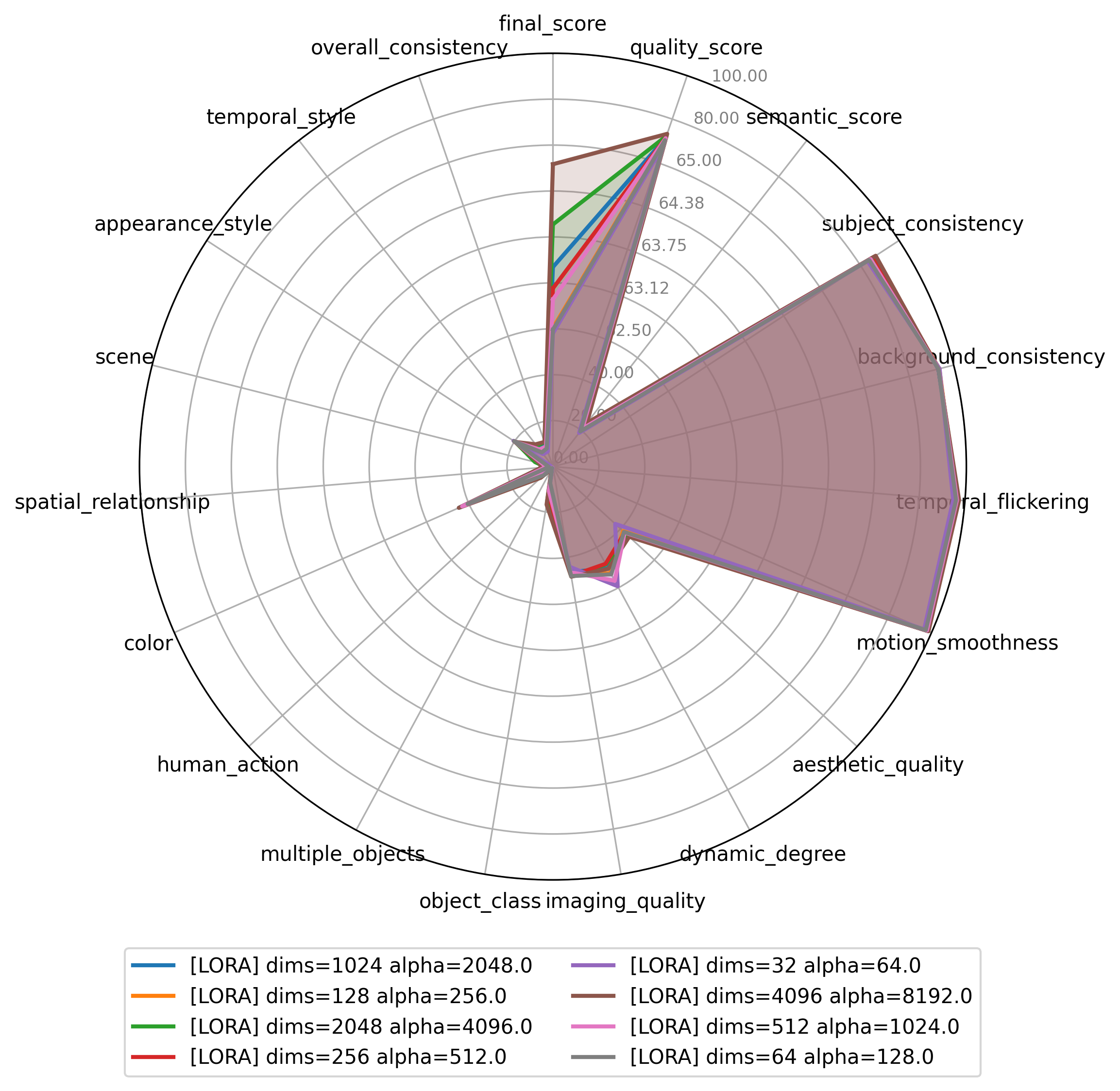}
    \caption{}
    \label{fig:text_enc_abl_lora}
  \end{subfigure}

  \caption{\textbf{Ablations for Text-Encoder Distillation.} We ablate the loss weights $w_\text{mse}$ and $w_\text{cd}$ for the \textbf{[RM]} mode in \textbf{(a)}; and ablate the two controllable hyperparameters of the LoRA layers, namely dimensions (\texttt{dims}) and the scale (\texttt{alpha}) of \textbf{[LORA]} mode in \textbf{(b)}.}
  \label{fig:tex_enc_abl_1}
\end{figure}

Table~\ref{tab:te_quant_main} presents a detailed quantitative evaluation of our proposed Text-Encoder Distillation framework, comparing multiple configurations of the distilled encoder and its associated $\mathit{CA}$ module. The baseline configuration, which employs the original \tfxxl encoder, achieves a VBench Total score of 80.31, establishing the upper bound for performance within our setup. Remarkably, when \tfxxl is replaced with the significantly smaller \dtf encoder paired with a 4-layer MLP-based $\mathit{CA}$ operating in Replace-Mode, the system maintains a high VBench Total score of \textbf{79.64} (see fig.~\ref{fig:text-encoder-qualitative}). This reflects a minimal performance drop of just 0.67 points, while delivering substantial reductions in parameter count and computational overhead.

In comparison, the Extend-Mode configuration—where the new $\mathit{CA}$ augments rather than replaces the original $\mathit{CE}$—incurs a slightly higher parameter overhead due to the dual-module setup. Interestingly, this configuration yields a marginally lower VBench score than Replace-Mode, which we attribute to the rigidity of the frozen $\mathit{CE}$ embedding space, potentially limiting the adaptability of the extended context representation.

\begin{table}[!t]
\centering
\caption{\textbf{Quantitative Evaluation of Text-Encoder Distillation}. \textbf{\#Parameters} ($\downarrow$) and \textbf{Vbench}($\uparrow$) scores are reported for different combinations of trainable \textbf{[\tdtf]} or frozen \dtf paired with $\mathit{CA}$ applied in \textbf{[RM]} Replace-Mode or \textbf{[EM]} Extend-Mode and using a 4-layer MLP v/s \textbf{[LORA]} LoRA layers as the $\mathit{CA}$.} 
\label{tab:te_quant_main}
\begin{tabular}{|l|c|c|c|c|}
\hline
\textbf{Method} & \textbf{\#Parameters ($\downarrow$)} & \multicolumn{3}{c|}{\textbf{VBench Score}} \\
\cline{3-5}
 & \textbf{(TE+CA)} & \textbf{Tot.($\uparrow$)} & \textbf{Qual.($\uparrow$)} & \textbf{Sem.($\uparrow$)} \\
\hline
\cellcolor{tabbaseline}\tfxxl Baseline & \cellcolor{tabbaseline} 4.732 B & \cellcolor{tabbaseline} 80.31 & \cellcolor{tabbaseline} 83.68 & \cellcolor{tabbaseline} 66.81 \\
\hline
\dtf $\mathit{CA}$ \textbf{[RM]} Replace-Mode & \cellcolor{tabsecond} 0.260 B & \cellcolor{tabfirst} 79.64 & \cellcolor{tabfirst} 83.71 & \cellcolor{tabfirst} 63.39 \\
\hline
\dtf $\mathit{CA}$ \textbf{[EM]} Extend-Mode & \cellcolor{tabthird} 0.266 B & \cellcolor{tabthird} 79.16 & \cellcolor{tabsecond} 83.56 & \cellcolor{tabthird} 61.55 \\
\hline
\dtf $\mathit{CA}$ \textbf{[LORA]} LoRA Mode & \cellcolor{tabfirst} 0.136 B & 64.74 & 74.94 & 24.08 \\
\hline
\textbf{[\tdtf]} Trainable \dtf Mode & \cellcolor{tabfirst}  0.136 B & \cellcolor{tabsecond} 79.20 & \cellcolor{tabthird} 83.44 & \cellcolor{tabsecond} 62.12 \\
\hline
\end{tabular}
\end{table}

For the \textbf{[LORA]} variant, we observe that achieving even minimum viable video generation quality requires a significantly increased number of low-rank dimensions and a high scalar \texttt{alpha} (see fig.~\ref{fig:text_enc_abl_lora}). Even with these adjustments, the resulting VBench score of 64.47, while enabling some visual fidelity, falls short of the performance achieved by other configurations. This motivated further experimentation with a trainable \dtf encoder \textbf{[\tdtf]}. Notably, this setup achieves a strong VBench score of \textbf{79.20} with only half the parameter count of the Replace-Mode configuration. However, we ultimately select \textbf{[RM]} as our final deployment choice, prioritising even marginal gains in VBench score to maximise generation quality, despite the higher parameter count relative to the \textbf{[\tdtf]} variant. Figure~\ref{fig:text-encoder-qualitative} presents qualitative examples that align with our experimental findings. Notably, the \textbf{[LORA]} mode frequently overlooks key semantic cues from the text prompts—for instance, generating a teddy bear instead of a panda, or failing to adhere to the black-and-white constraint specified in the prompt. In contrast, the remaining three modes perform comparably, with only occasional semantic mismatches (Darth Vader instead of Yoda).

In Figure~\ref{fig:tex_enc_abl_1}, we present a couple of ablations of our proposed Distillation Framework. Chosen for its superior empirical performance, a natural question regarding \textbf{[RM]} that arises, is how the two distinct loss functions in Equation~\ref{eq:text_encoder_loss} contribute to the overall optimisation landscape. Figure~\ref{fig:tex_enc_abl_lossweight} illustrates various combinations of weights applied to the sum of these losses. When either $w_\text{mse}$ or $w_\text{cd}$ is set to zero, the corresponding loss is effectively disabled. We note that disabling the Cosine Distance loss results in a divergence, as denoted by the purple curve in Figure~\ref{fig:tex_enc_abl_lossweight} (observe the center of the radar-plot). This underscores that the Cosine Distance loss is essential for stabilising the training. Lastly, as can be observed, the best performance is obtained when we use $w_\text{mse}=1.0$ and $w_\text{cd}=0.1$.

Next, following the analysis of loss weighting, we investigate how the architecture of the ContextAdapter influences the adaptability of the distilled text encoder \dtf. A comprehensive exploration of the architectural design space is beyond the scope of this work, so we ablate the number of LoRA dimensions (\texttt{dims}) and the LoRA scale (\texttt{alpha}) in Figure~\ref{fig:text_enc_abl_lora} for the \textbf{[LORA]} mode. Our findings indicate that even minimal visual quality in this mode necessitates a High-Rank Approximation rather than a Low-Rank one. This observation further motivates the need for full fine-tuning of the \dtf model’s weights, which we address in the \textbf{[\tdtf]} mode.

Overall, these results highlight the effectiveness of our distillation strategy. The distilled encoder—trained solely on generative text prompts without any image or video supervision—retains the semantic and perceptual quality of generated videos to a level nearly indistinguishable from the original large-scale model. Despite minimal performance degradation, the approach delivers substantial efficiency gains, making it highly suitable for deployment on resource-constrained mobile hardware. These findings confirm our hypothesis that the full capacity of \tfxxl is not required for high-quality video synthesis; thereby, a carefully distilled encoder can serve as a viable drop-in replacement without compromising user experience or output fidelity. The \textbf{[RM]} configuration is integrated into the final end-to-end \name pipeline, as shown in Figure~\ref{fig:final_full_pipeline_illustration}. Notably, while the original \tfxxl text encoder was infeasible for on-device execution, the distilled version achieves a remarkable latency of \textbf{3ms} on the \platform~packaged into the \laptopsoc~platform, enabling real-time performance.

\subsection{Asymmetric Decoder Distillation}
\label{subsec:asymm_dec_distil}
\begin{figure}[!t]
  \centering
  \includegraphics[width=0.9\textwidth]{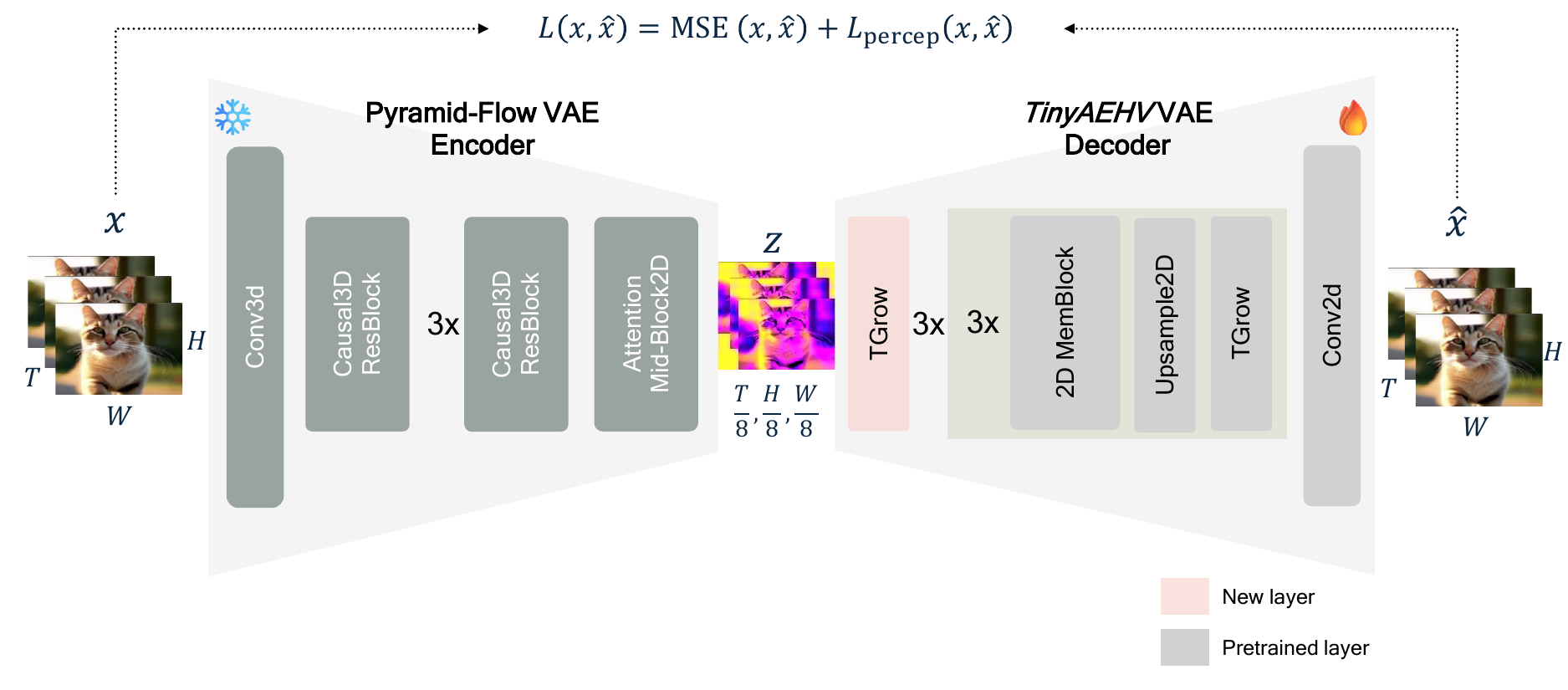}
  \captionof{figure}{\textbf{Overview of the proposed Asymmetric Decoder Distillation framework.} A new decoder from a different pretrained latent video diffusion model is distilled into our pipeline by: firstly modifying the decoder architecture to match the fixed \texttt{[8$\times$8$\times$8]} compressed latent-space of our model; and secondly by finetuning this asymmetric VAE with video data using MSE and LPIPS \cite{zhang2018unreasonable} losses. The encoder is kept frozen so that the generative latent-space of the video diffusion backbone is undisturbed. We note that the \textit{TinyAEHV} \cite{BoerBohan2025TAEHV} decoder is visualised here, but the framework works with other models as well.}
  \label{fig:as_dec_dist}
\end{figure}

Having addressed the challenge of the large Text-Encoder for mobile video generation, we proceeded to port the model to the device, only to encounter a second bottleneck. Although the native codec-latent-decoder in the base model is relatively lightweight in terms of parameters (226M), its forward computation graph requires storing large 4D feature-map buffers. This made it impossible to fit even a single forward pass of the decoder on the mobile platform. Using a smaller latent representation of [\texttt{7×10×16}] allowed the graph to fit on \laptopsoc, but the execution time remained prohibitively high (\textbf{3500ms}). A deeper analysis of the slow operations revealed that the conv3D operation is the primary bottleneck—an operation that is indispensable for causal video auto-encoding.

Rather than designing a new \textit{mobile-friendly} codec-latent-VAE from scratch—which would require prohibitively large amounts of data and compute—we frame this challenge as a distillation problem. Existing open-source video generation models \cite{yang2024cogvideox, jin2024pyramidal, hunyuan2025, HaCohen2024LTXVideo, wan2025, lin2024opensora} each employ their own codec-latent-VAEs, resulting in diverse video latent spaces for diffusion. We hypothesize that the decoder from one of these models may be sufficiently efficient to serve as a mobile-friendly candidate. Something that we can distill into our pipeline and enable on-device execution. This hypothesis raises two key questions: (i) \emph{Are the video latent spaces across different models easily transferable (through lightweight finetuning)?} and (ii) \emph{How can we reconcile disparities in latent compression factors among these models?} To address both questions within a unified empirical framework, we propose an Asymmetric Decoder Distillation strategy, detailed as follows.

Our proposed framework comprises \textbf{three components} (see fig.~\ref{fig:as_dec_dist}). \textbf{First}, we introduce asymmetry into the codec-latent-VAE by retaining the original encoder, $\mathcal{E}_{\text{enc}}$, to produce coded latents $\vz = \mathcal{E}_{\text{enc}}(\vx)$, while replacing the original decoder with $\mathcal{E}_{\text{dec}}$ a new one, $\mathcal{F}_{\text{dec}}$, to reconstruct videos as $\hat{\vx} = \mathcal{F}_{\text{dec}}(\vz)$. Since $\mathcal{F}_{\text{dec}}$ was originally trained for a different latent space, fine-tuning is essential. However, before fine-tuning, we must resolve the mismatch in compression factors between the base encoder and the asymmetric decoder. \textbf{Second}, we minimally adapt the decoder architecture to match the fixed encoder’s compression factor of \texttt{[8$\times$8$\times$8]}. This adjustment involves either adding or removing blocks, depending on the decoder’s original compression ratio. When new blocks are introduced, we reuse the existing architectural design as much as possible and minimise additional parameters. \textbf{Finally third}, we fine-tune the entire setup end-to-end using a reconstruction objective, $\mathcal{L}(\vx, \hat{\vx})$, combining MSE and LPIPS losses~\cite{zhang2018unreasonable}. The encoder remains frozen to preserve the latent space required by MMDiT, which also allows us to omit the KL regularizer typically employed in VAE training.

\begin{table}[!t]
\centering
\caption{\textbf{Quantitative Evaluation of Asymmetric Decoder Distillation}. We report the DAVIS \cite{pont20172017} \textbf{PSNR} ($\uparrow$) using the original Encoder (without modification and finetuning), and using our pipeline's Encoder (after distillation), and \textbf{VBench} scores ($\uparrow$) for evaluating the reconstruction performance and the generative decoding performance respectively.} 
\label{tab:asymm_dec_quant_main}
\resizebox{\textwidth}{!}{%
\begin{tabular}{|l|c|c|c|c|c|c|c|}
\hline
\textbf{Method} & \textbf{\#Params ($\downarrow$)} & \textbf{Nvidia} & \textbf{DAVIS} & \textbf{DAVIS} & \multicolumn{3}{c|}{\textbf{VBench Score}} \\
\cline{6-8}
 & \textbf{Decoder} & \textbf{H100} & \textbf{PSNR ($\uparrow$)} & \textbf{PSNR ($\uparrow$)} & \textbf{Tot.($\uparrow$)} & \textbf{Qual.($\uparrow$)} & \textbf{Sem.($\uparrow$)} \\
 &  & \textbf{Lat. ($\downarrow$)} & \textbf{Orig. Enc.} & \textbf{Our Enc.} &  &  &  \\
\hline
\cellcolor{tabbaseline}\baseline~Native Decoder & \cellcolor{tabbaseline} 226M & \cellcolor{tabbaseline} 2.496s & \cellcolor{tabbaseline}29.12 & \cellcolor{tabbaseline}29.12 & \cellcolor{tabbaseline}80.31 & \cellcolor{tabbaseline}83.68 & \cellcolor{tabbaseline}66.81 \\
\hline
WAN modified &\cellcolor{tabthird}74M &\cellcolor{tabthird}1.666s &\cellcolor{tabfirst} 31.47 &\cellcolor{tabthird}29.18 &\cellcolor{tabfirst}80.36 &\cellcolor{tabfirst}83.82 &\cellcolor{tabsecond}66.55 \\
\hline
Cosmos CV\texttt{[8x8x8]} & \cellcolor{tabsecond} 63M & \cellcolor{tabfirst} 0.451s &\cellcolor{tabsecond}29.34 &\cellcolor{tabsecond}29.45 & 79.96 & 83.37 & \cellcolor{tabthird}66.35 \\
\hline
LTXVideo modified & 237M & 1.738s &\cellcolor{tabthird}28.34 &\cellcolor{tabfirst}29.46 & \cellcolor{tabsecond}80.34 & \cellcolor{tabsecond}83.75 & \cellcolor{tabfirst}66.68 \\
\hline
(Our) TinyAEHV modified & \cellcolor{tabfirst} 10M & \cellcolor{tabsecond} 0.851s & 27.71 & 28.40 & \cellcolor{tabthird}80.25 & \cellcolor{tabthird}83.51 & 67.19 \\
\hline
\end{tabular}
}
\end{table}

As shown in Figure~\ref{fig:as_dec_dist}, we apply minimal modifications to integrate different asymmetric decoders into our pipeline. For the TinyAEHV decoder~\cite{BoerBohan2025TAEHV}, we modify the first \texttt{TGrow} (temporal upsampler) layer to perform $2\times$ temporal upsampling instead of its default $1\times$ (no) upsampling, reinitialising the parameters of this block with random weights. This single change suffices to match our latent compression factor. For the Cosmos decoder~\cite{kwon2019cosmos}, we use the Continuous Tokens variant with [\texttt{8$\times$8$\times$8}] compression, requiring no architectural changes. For LTXVideo~\cite{HaCohen2024LTXVideo}, we remove the decoder’s unpatchification layer and update the \texttt{conv\_out} layer with new weights. Additionally, to accommodate our 16-dimensional latents, we replace the \texttt{conv\_in} layer. Finally, for the Wan decoder~\cite{wan2025}, similar to TinyAEHV, we modify the first upsampling block to perform $2\times$ spatio-temporal upsampling instead of the default $2\times$ spatial-only upsampling. These minimal adjustments enable us to distill diverse asymmetric decoders into our generation pipeline.

This setup was trained using the AdamW optimiser with a fixed learning rate of 1e-4. Compared to the earlier Text-Encoder Distillation experiments (ref subsec.~\ref{subsec:text_enc_distil}), these runs were significantly more GPU-intensive. Training was performed for 200{,}000 iterations on eight 80\,GB H100 GPUs, with per-GPU batch sizes ranging from 2 to 6 (depending on decoder size), resulting in an effective batch size of 16--48. We used the default patch size of \texttt{[33$\times$256$\times$256]} sampled from a corpus of approximately $\sim$350K videos. The full training process took about $\sim$120-140 hours. For the reconstruction objective, we followed \baseline's default weighting: $10.0$ for the MSE loss and $1.0$ for the LPIPS loss~\cite{zhang2018unreasonable}.

Table~\ref{tab:asymm_dec_quant_main} summarises our experiments with the proposed Asymmetric Decoder Distillation framework. Remarkably, even with minimal architectural modifications, all decoder variants perform well. The PSNR scores on the DAVIS~\cite{pont20172017} test set average above \textbf{29dB}, indicating that the asymmetric latent VAE can faithfully reconstruct video signals while operating through the frozen generative latent space of MMDiT. Although these results are preliminary, they provide strong empirical evidence for the universal nature of compressive video latent spaces learned by different models, demonstrating that such spaces can be transferred between each other with relatively low fine-tuning cost.

For our deployment, the TinyAEHV decoder~\cite{BoerBohan2025TAEHV} proved to be the most parameter-efficient and mobile-friendly option. While the native decoder could not run on \platform~for profiling, the modified version achieves a latency of \textbf{143ms} when decoding a \texttt{[49$\times$320$\times$512]} video from a latent tensor of shape \texttt{[7$\times$40$\times$64]} on the \platform. This distilled decoder is integrated into our final optimised \name pipeline (see fig.~\ref{fig:final_full_pipeline_illustration}).

\subsection{MMDiT Block-Pruning}
\label{subsec:block_pruning}

\begin{figure}[!t]
  \centering
  \includegraphics[width=0.755\textwidth]{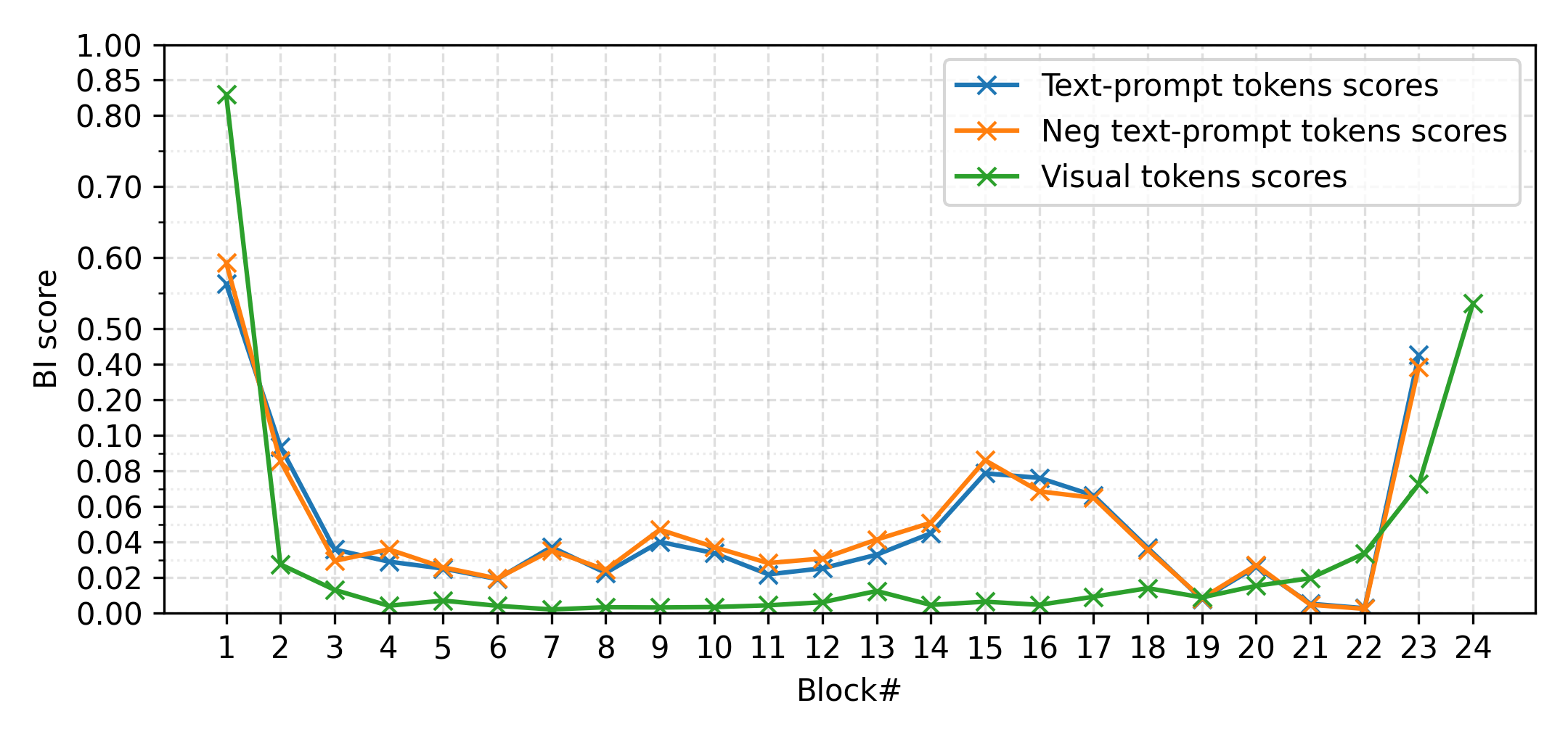}
  \captionof{figure}{\textbf{Block Importance Scores v/s Block-ids.} Block Importance Scores for the 24 MMDiT blocks in the denoiser backbone, calculated using equation \ref{eq:bi_scores}. Textual scores are computed for 23 blocks, excluding the final block where text tokens are ignored. The plot visualises token-level importance scores across two CFG forward passes: blue for descriptive text-prompts and orange for negative text-prompts. Visual token scores (green) are shown only once, as they remain identical across both passes.}
  \label{fig:block_pruning_biscores}
\end{figure}

\begin{figure}[!t]
  \centering
  \includegraphics[width=\textwidth]{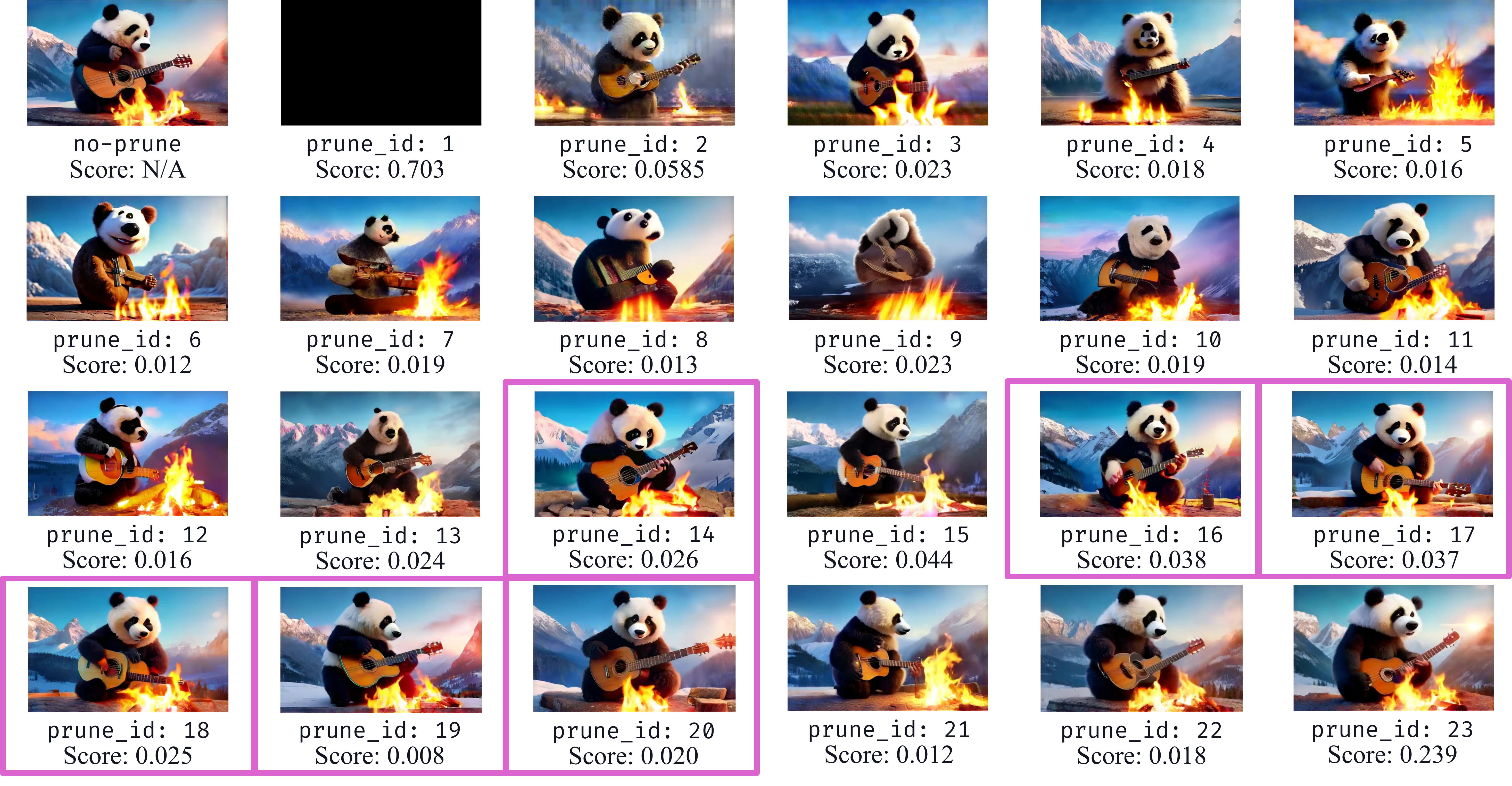}
  \captionof{figure}{\textbf{Visual guidance for Block Pruning.} We visualise randomly selected frames from the generated \texttt{[49×320×512]} videos, across 24 different models in which the \texttt{prune\_id}$^\text{th}$ MMDiT block is dropped from the model. We choose the 6 blocks highlighted with the boxes for pruning, giving us 25\% model size reduction.}
  \label{fig:block_pruning_visguid}
\end{figure}

After addressing the two major challenges—the oversized Text-Encoder and the unoptimised Decooder—we successfully obtained our first end-to-end \platform~latency measurement of \textbf{$\sim$184.2s} on the \laptopsoc~soc. While this result demonstrates the feasibility of mobile video generation, the total runtime of approximately three minutes to produce a 2-second video at a relatively low resolution [\texttt{320$\times$512}] is far from our goal. Interestingly, the Text-Encoder and the Decoder, which initially prevented on-device execution, account for only 0.2s of the total latency. The remaining 184s are required for the \textit{spatio-temporally pyramidal causal} latent generation performed by the MMDiT denoiser. We can only imagine how much longer a monolithic fully-bidirectionally attentive transformer like Wan~\cite{wan2025} would require for generating the same sized latent-video. This observation motivates two key optimisation directions: (i) reducing the size of MMDiT without compromising quality, thereby accelerating each denoising step; and (ii) reducing the number of denoising iterations (NFEs) required to generate the latent video. In this subsection, we focus on the first direction, while the second direction is explored in Subsection~\ref{subsec:step_distil}.

The MMDiT architecture, introduced in Stable Diffusion 3~\cite{esser2024scaling}, extends the original Diffusion Transformer (DiT) \cite{peebles2022dit} into a \textit{Multi-Modal} variant. MMDiT enhances the expressiveness of the Transformer by allowing text tokens to attend to visual tokens, thereby influencing and updating their representations through the model’s layers. Despite this multi-modal design, the architecture remains a stack of residual blocks applied without spatial or temporal down/up-sampling of token maps, unlike earlier UNet based designs. This structure presents two main optimisation strategies: (i) pruning entire residual blocks~\cite{fang2025tinyfusion,xie2025sana,sreenivas2024llm}, or (ii) performing fine-grained pruning within blocks by removing unused layers or operations (width pruning)~\cite{wu2025taming}. Building on insights from TinyFusion~\cite{fang2025tinyfusion}, which reports superior speed-ups and compression ratios for Block Pruning compared to Width Pruning, we prioritise Block Pruning. This choice is further motivated by hardware considerations: \platform~supports static, repetitive compute graphs more efficiently than asymmetric or conditional compute graphs, and Block Pruning is generally more quantisation-friendly. For these reasons, we adopt Block Pruning as our primary strategy. 

To this end, we propose a block-pruning strategy inspired by SANA-1.5 \cite{xie2025sana}, but adapted to the MMDiT architecture and extended with a full-teacher fine-tuning stage. The approach begins by analysing the relative importance of MMDiT backbone blocks and pruning the least important ones. This is followed by data-driven fine-tuning of the pruned model, which we refer to as Stage-1 fine-tuning. Finally, we perform an additional fine-tuning stage using the full teacher model for further alignment, which we refer to as the Stage-2 fine-tuning.

\begin{figure}[!t]
  \centering
  \includegraphics[width=0.7\textwidth]{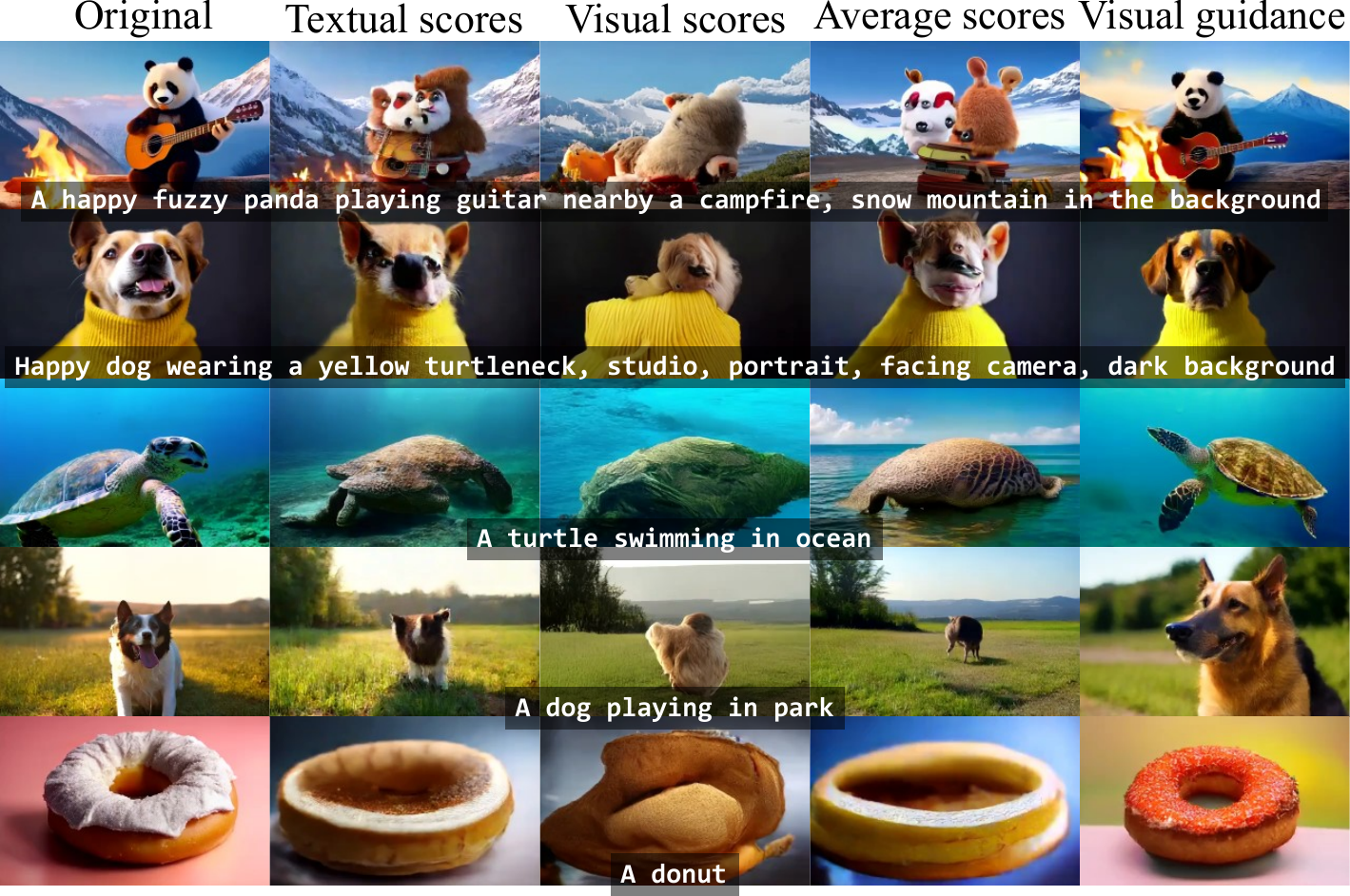}
  \captionof{figure}{\textbf{Different BI-scores based block-pruning compared to visual guidance.} We visualise randomly selected frames from the
generated \texttt{[49×320×512]} videos corresponding to the adjacent text prompts, across different 18-blocks pruned versions of the original model. Using only the textual-scores, or visual-scores or even average of the two (see fig. \ref{fig:block_pruning_biscores}) directly results in a lot of semantic distortion in the generated samples after Stage-1 finetuning. Whereas, upon choosing the blocks to prune based on the average scores as well as the visual impact (see fig. \ref{fig:block_pruning_visguid}), causes minimal semantic distortion.}
  \label{fig:block_pruning_scorevisqualeffect}
\end{figure}

\paragraph{Analysing Block Importance.}
The MMDiT denoiser, denoted as $\dit$, can be expressed as a composition of $N$ blocks operating on tokens concatenated from three sources: visual latent tokens $\vz$ (diffused with noise), textual tokens $\hat{t}$ derived from the \texttt{prompt} (ref eq.~\ref{eq:text_encoder_def}), and the clip embedding token $\hat{c}$. \baseline trains the denoiser starting from StableDiffusion-3.5~\cite{esser2024scaling} MMDiT checkpoint, hence here $N = 24$.
\begin{align}
    &\dit(\vz, \hat{t}, \hat{c}) = \dit_\mathit{N} \circ \dit_\mathit{N-1} \circ \dit_\mathit{k} ... \circ \dit_\mathit{1}(\vz, \hat{t}, \hat{c}) \\
    &\text{where, } \hat{c} := \mathit{CLIP}(\text{\texttt{prompt}})
\end{align}
Due to the multi-modal nature of the MMDiT architecture, we obtain separate block importance scores for the visual and the textual tokens, represented by $\mathit{BI}^v_k$ and $\mathit{BI}^v_k$ respectively, for the $k^\text{th}$ block. 
\begin{align}
    &\vz_\mathit{k + 1}, \hat{t}_\mathit{k + 1} = \dit_k(\vz_\mathit{k}, \hat{t}_\mathit{k}, \hat{c}) \\
    &\mathit{BI}^v_k := 1 - \mathbb{E}\left[ \frac{ z_k . z_\mathit{k + 1}}{\lVert z_k \rVert_2 \lVert z_\mathit{k + 1} \rVert_2 } \right] 
    \text{ and  }
    \mathit{BI}^t_k := 1 - \mathbb{E}\left[ \frac{ \hat{t}_k . \hat{t}_\mathit{k + 1}}{\lVert \hat{t}_k \rVert_2 \lVert \hat{t}_\mathit{k + 1} \rVert_2 } \right] \\
    &\mathit{BI}_k := \left(\mathit{BI}^v_k, \mathit{BI}^t_k \right) \label{eq:bi_scores}
\end{align}

As shown in Figure~\ref{fig:block_pruning_biscores}, we compute Block Importance scores for each \(k^\text{th}\) block in the MMDiT (see Eq.~\ref{eq:bi_scores}), defined as the Cosine Distance between the block $\mathcal{D}_k$’s input and output tokens. To estimate these scores, we use a small but diverse calibration set of 100 text prompts, generating five sample videos for each. During this process, we probe the internal token representations of the MMDiT \(\mathcal{D}\) at every denoising step for both CFG (Classifier-Free Guidance~\cite{ho2022classifier}) passes—one with descriptive prompts and one with negative prompts. Consistent with observations for SANA-1.5~\cite{xie2025sana}, we find that the initial and final blocks are more influential, while intermediate blocks contribute less, as they induce minimal residual changes to the tokens. Interestingly, due to the model’s multi-modal nature, the visual and textual importance of a block are not correlated. Therefore, as illustrated in Figure~\ref{fig:block_pruning_visguid}, we also assess the impact of removing each block on the final generation quality. Based on both the importance scores and visual impact, we select six highlighted blocks for pruning, while also experimenting with smaller and slightly larger sets to explore the trade-off between quality and model size.

\begin{table}[!t]
\centering
\caption{\textbf{Quantitative Evaluation of MMDiT Block-Pruning}. 
Performance of the proposed MMDiT Block-Pruning strategy across different model sizes, reported using \textbf{VBench scores} ($\uparrow$) after Stage-1 and Stage-2 fine-tuning. For each configuration, we show model size (\textbf{\#Parameters}, $\downarrow$) and Qualcomm \platform~\textbf{latency} ($\downarrow$). Based on this trade-off, we select the 18-block variant for the final end-to-end pipeline (see Fig.~\ref{fig:final_full_pipeline_illustration}). Latency is measured as the sum of one denoising step across all three pyramidal stages (resolutions). To reduce the cost of latency profiling, measurements are reported only for the base model and the selected 18-block variant.}
\label{tab:bp_quant_main}
\begin{tabular}{|c|l|c|c|c|c|c|}
\hline
\textbf{Stage} &\textbf{Method} & \textbf{\#Params ($\downarrow$)} & \textbf{Qualcomm} & \multicolumn{3}{c|}{\textbf{VBench Score}} \\
 &  & \textbf{MMDiT}  & \textbf{Hexagon NPU} & \multicolumn{3}{c|}{} \\
\cline{5-7}
 &  &  &  \textbf{Latency ($\downarrow$)} & \textbf{Tot.($\uparrow$)} & \textbf{Qual.($\uparrow$)} & \textbf{Sem.($\uparrow$)} \\
\hline

\textbf{Baseline} & \cellcolor{tabbaseline}24 Blocks MMDiT & \cellcolor{tabbaseline} 2.028 B & \cellcolor{tabbaseline} 1.15s & \cellcolor{tabbaseline} 80.31 & \cellcolor{tabbaseline} 83.68 & \cellcolor{tabbaseline} 66.81 \\

\hline

& 22 Blocks MMDiT & 1.858 B & - & \cellcolor{tabfirst} 79.82 & \cellcolor{tabfirst} 83.30 & \cellcolor{tabfirst} 65.92 \\

\textbf{1} & 20 Blocks MMDiT & \cellcolor{tabthird} 1.688 B & - & \cellcolor{tabsecond} 78.65 & \cellcolor{tabsecond} 82.36 & \cellcolor{tabthird} 63.82 \\

& 18 Blocks MMDiT & \cellcolor{tabsecond} 1.518 B &\cellcolor{tabfirst}0.74s & \cellcolor{tabthird} 78.39 & \cellcolor{tabthird} 81.58 & \cellcolor{tabsecond} 65.63 \\

& 16 Blocks MMDiT & \cellcolor{tabfirst} 1.348 B & - &  74.59 &  78.74 & 57.99 \\

\hline

\textbf{2} & 18 Blocks MMDiT &\cellcolor{tabsecond} 1.518 B &\cellcolor{tabfirst}0.74s & \cellcolor{tabfirst} 80.21 & \cellcolor{tabfirst} 83.54 & \cellcolor{tabfirst} 66.90 \\

& 16 Blocks MMDiT &\cellcolor{tabfirst} 1.348 B & - & \cellcolor{tabsecond} 78.62 & \cellcolor{tabsecond} 82.40 & \cellcolor{tabsecond} 63.50 \\

\hline

\end{tabular}
\end{table}
\begin{figure}[!t]
  \centering
  \includegraphics[width=\textwidth]{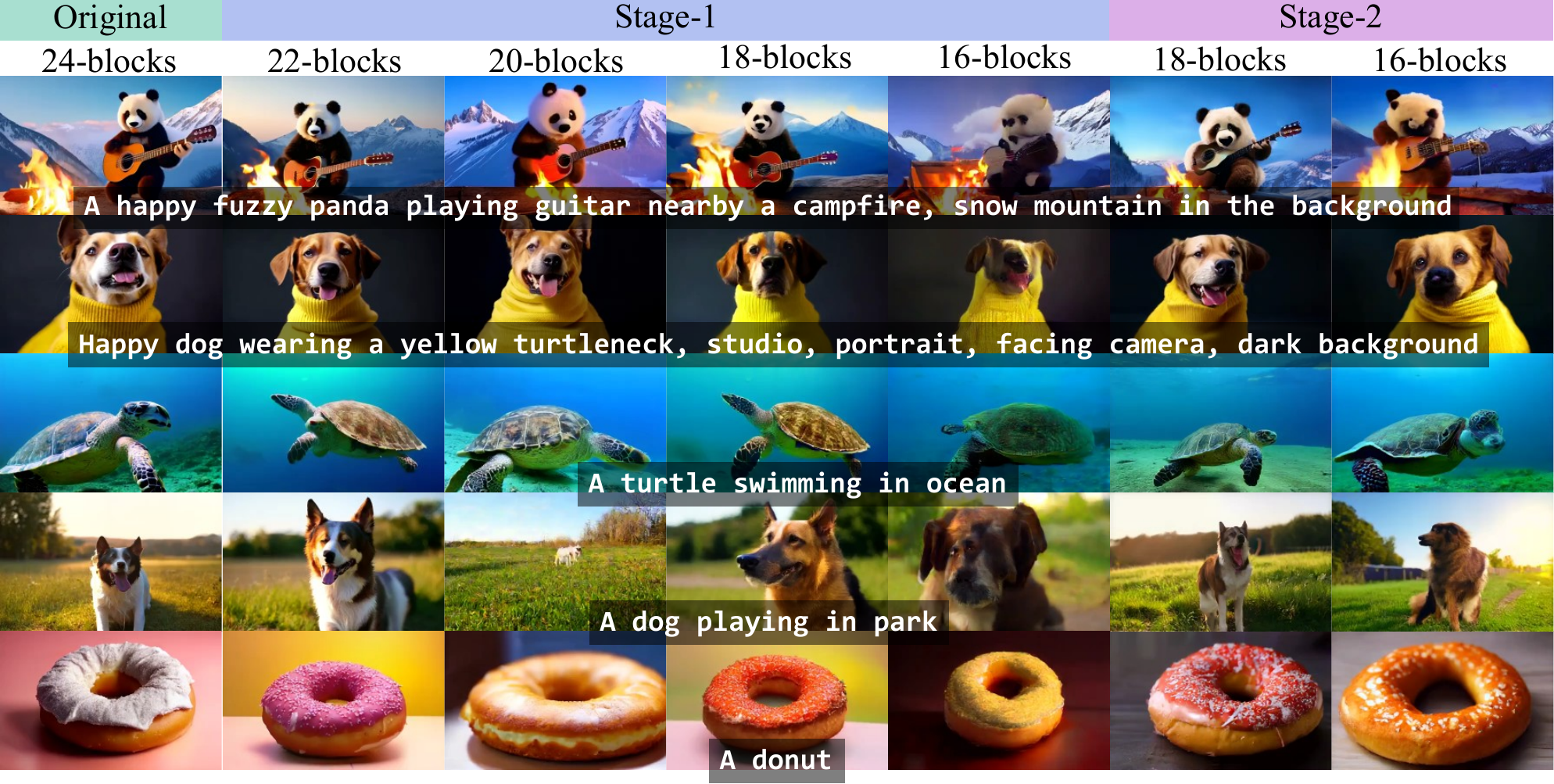}
  \captionof{figure}{\textbf{Qualitative Evaluation of MMDiT Block Pruning.} We visualise randomly selected frames from the generated \texttt{[49×320×512]} videos, across models with different pruned-sizes, after ground truth video data based Stage-1 finetuning, and Full Teacher model distillation based Stage-2 finetuning.}
  \label{fig:block_pruning_qual_main}
\end{figure}

\paragraph{Stage-1 Finetuning.}
After removing the selected blocks from the MMDiT, we finetune the pruned model using ground-truth data. Specifically, we sample training batches from our curated dataset of approximately $\sim$350K videos and their corresponding prompts. Fine-tuning is performed with the original Flow-Matching objective~\cite{lipman2022flow}. We adopt the default setting of \baseline~\cite{jin2024pyramidal}, where only the current frame is denoised while conditioning on past frames sampled from ground truth. To improve robustness to test-time generations, these history frames are corrupted with gaussian noise during training.

Table~\ref{tab:bp_quant_main} summarises the performance of block-pruned models of different sizes after Stage-1 fine-tuning. Based on these results, we select the 18-block model for our final pipeline, as it offers the best trade-off between model size and generation quality. Figure~\ref{fig:block_pruning_scorevisqualeffect} illustrates the impact of selecting six blocks for pruning (out of 24, resulting in an 18-block model) using only block-importance scores versus incorporating visual impact in the selection process. While this step remains manual in our approach, we note that automating visual guidance using LPIPS~\cite{zhang2018unreasonable} or other specialised networks is an interesting direction for future work.

This setup was trained using the Adam optimiser with a fixed learning rate of 3e-5 on four 80GB NVIDIA H100 GPUs, with a per-GPU batch size of 4, resulting in an effective batch size of 16. Since training was limited to 300 iterations, the process took only about 1--2 hours. Although we experimented with longer training (up to 3K iterations), we observed no significant performance gains. Remarkably, even with such minimal fine-tuning, we were able to recover most of the lost performance, underscoring the effectiveness of our block selection strategy based on importance scores and visual inspection.

\begin{figure}[!t]
  \centering
  
  \begin{subfigure}[t]{0.48\textwidth}
    \centering
    \includegraphics[width=\textwidth]{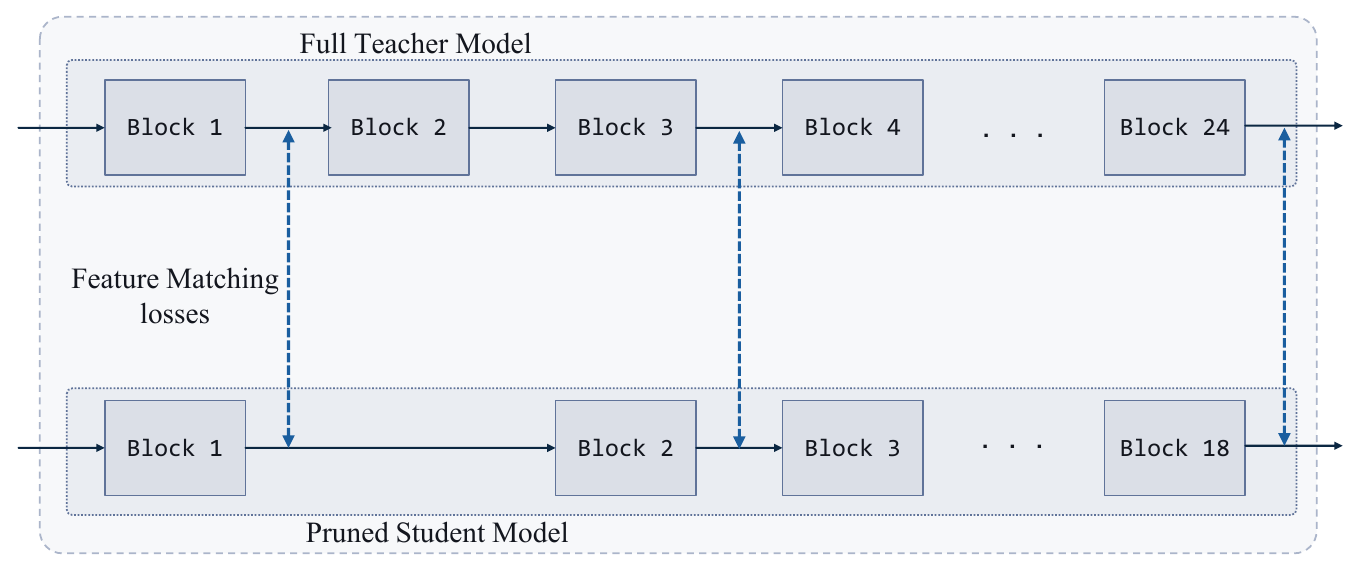}
    \caption{Simple Block Mapping}
    \label{fig:block_pruning_simpleblkmp}
  \end{subfigure}
  \hfill
  \begin{subfigure}[t]{0.48\textwidth}
    \centering
    \includegraphics[width=\textwidth]{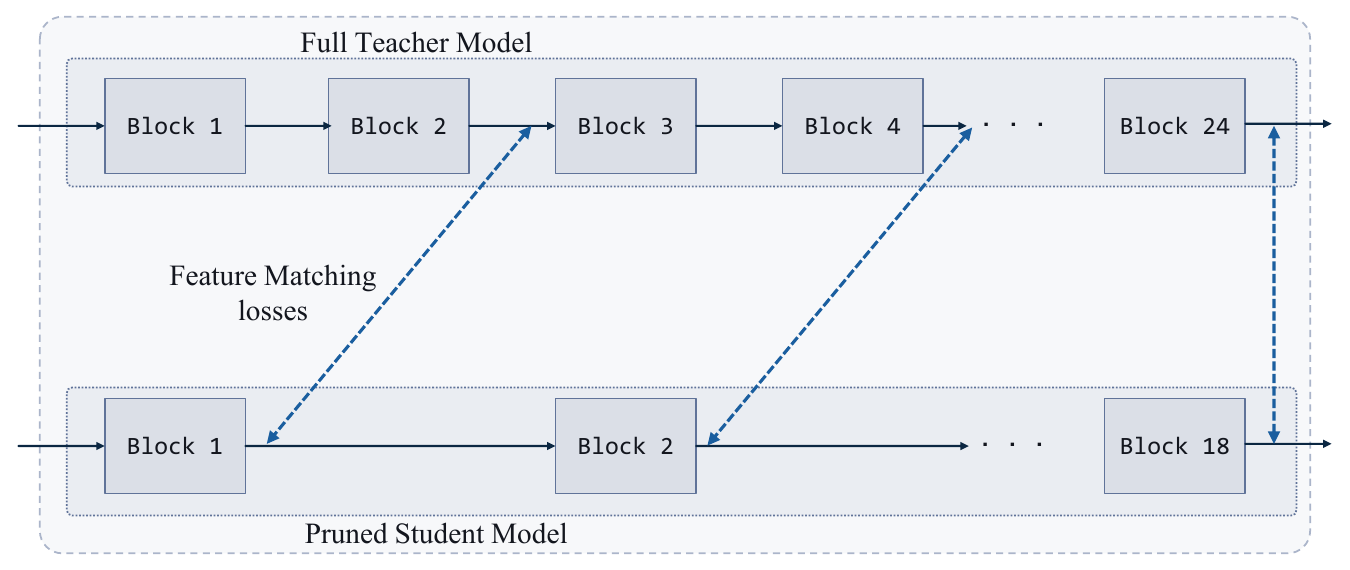}
    \caption{Next-Block Mapping}
    \label{fig:block_pruning_nextblkmp}
  \end{subfigure}

  \caption{\textbf{Stage-2 Simple Block Mapping v/s Next-Block Mapping.} For minimising the token-matching losses between the Full Teacher model and the Pruned Student model, \textbf{(a)} Simple Block Mapping maps the output of each of the present block in the student model to the corresponding one in the Teacher model; whereas, \textbf{(b)} Next-Block Mapping maps the output of each present block in the Student model to the input of the next-available block in the Teacher model (except for the final block which always matches final output). }
  \label{fig:block_pruning_blkmp}
\end{figure}
\begin{table}[!t]
\centering
\caption{\textbf{MMDiT Block-Pruning Stage-2 Ablations}. We ablate the choices over: the three losses, namely Feature Loss (token-matching), Teacher FM Loss (flow-matching using Teacher's predicted flow), and Data FM Loss (flow-matching using ground truth data flow); and which Block-Mapping to use when Feature Loss is active using \textbf{VBench} scores ($\uparrow$).} 
\label{tab:bp_quant_abls}
\begin{tabular}{|c|c|c|c|c|c|c|}

\hline
\textbf{Feature} & \textbf{Teacher} & \textbf{Data} & \textbf{Block} & \multicolumn{3}{c|}{\textbf{VBench Score}} \\
\cline{5-7}
\textbf{Loss} & \textbf{FM Loss} & \textbf{FM Loss} & \textbf{Mapping} & \textbf{Tot.($\uparrow$)} & \textbf{Qual.($\uparrow$)} & \textbf{Sem.($\uparrow$)} \\
\hline

\cellcolor{tabbaseline} & \cellcolor{tabbaseline} & \cellcolor{tabbaseline} \greentick & \cellcolor{tabbaseline} N/A & \cellcolor{tabbaseline} 78.39 & \cellcolor{tabbaseline} 81.58 & \cellcolor{tabbaseline} 65.63 \\

\hline
 & \greentick &  & N/A & 80.00 & \cellcolor{tabsecond} 83.52 & 65.89 \\

\greentick &  &  & Next-Block & 79.86 & 82.92 & \cellcolor{tabsecond} 67.61 \\

\greentick & \greentick &  & Next-Block & 79.93 & 82.90 & \cellcolor{tabfirst} 68.02 \\

 & \greentick & \greentick & N/A & \cellcolor{tabthird} 80.04 & \cellcolor{tabsecond} 83.52 & 66.11 \\

\greentick &  & \greentick & Next-Block & \cellcolor{tabfirst} 80.35 & \cellcolor{tabfirst} 83.82 & 66.48 \\

\greentick & \greentick & \greentick & Next-Block & \cellcolor{tabsecond} 80.11 & \cellcolor{tabthird} 83.44 & \cellcolor{tabthird} 66.79 \\
\hline

\greentick & \greentick &  & Simple & \cellcolor{tabsecond} 80.10 & \cellcolor{tabsecond} 83.31 & \cellcolor{tabfirst} 67.28 \\

\greentick & \greentick & \greentick & Simple & \cellcolor{tabfirst} 80.21 & \cellcolor{tabfirst} 83.54 & \cellcolor{tabsecond} 66.90 \\
\hline

\end{tabular}
\end{table}

\paragraph{Stage-2 Finetuning.} After establishing a lower bound on the performance achievable by the pruned models in Stage-1, we proceed with Stage-2 finetuning of the pruned model. In this stage, we incorporate feature-matching losses between the Full Teacher model and the pruned Student model. Specifically, we apply an MSE loss on visual tokens, a cosine distance loss on textual features, and Flow-Matching losses~\cite{lipman2022flow} using both the Teacher model's outputs and ground-truth flow from the data as supervision. Table~\ref{tab:bp_quant_main} reports the scores obtained after Stage-2 finetuning (second section). Notably, the 25\% pruned model with 18 blocks achieves a VBench score of \textbf{80.21}, only 0.1\% lower than the base 24-block model (\textbf{80.31}), enabling near-lossless compression of the MMDiT denoiser. Figure~\ref{fig:block_pruning_qual_main}, shows qualitative examples of the kind of generations that can be obtained by the differently sized pruned models after the Stage-1 and Stage-2 finetuning.

\emph{It is intriguing that Stage-2 finetuning does not perform well when applied directly to the pruned model, despite including the data-based Flow-Matching loss used in Stage-1. We also experimented with various annealed weighting schemes during training, but none matched the performance achieved by the curriculum approach of Stage-1 followed by Stage-2. We attribute this behaviour to the optimisation landscape induced by pruning, though a deeper investigation could provide valuable insights—an interesting direction for future work.}

Table~\ref{tab:bp_quant_abls} presents an ablation study on the design choices for loss functions used during Stage-2 finetuning. We also compare two block-mapping strategies: Simple Block mapping and Next-Block mapping, which determine how each block in the pruned Student model is paired with a block in the Full Teacher model for applying feature-matching losses. Figure~\ref{fig:block_pruning_blkmp} illustrates the difference between these two schemes. All experiments are conducted on the 18-block model. While all reasonable configurations perform well in principle, some yield slightly better results in practice. Interestingly, the model trained with Next-Block Mapping and only data-based Flow-Matching loss achieves the highest VBench score of \textbf{80.35} (even surpassing the base model). However, this configuration introduces artifacts in some generations and occasionally produces black videos for certain prompts. Therefore, we adopt the model trained with all losses and Simple Block Mapping as the final version for deployment in the \name~pipeline (see Fig.~\ref{fig:final_full_pipeline_illustration}). As reported in Table~\ref{tab:bp_quant_main}, starting from the full 24-block MMDiT with a \platform~latency of \textbf{1.15s}, we reduce the latency to \textbf{0.74s} with minimal impact on VBench performance.

\subsection{Step Distillation}
\label{subsec:step_distil}
Having pruned approximately 25\% of the MMDiT denoiser’s parameters, the video generation latency on \platform~decreased from \textbf{184.2s} to \textbf{118.6s}, yielding a saving of \textbf{65.8}, and thus, significantly improving time-to-video. Furthermore, we aim to reduce latency further to make the system more practical and user-friendly, with real-time generation remaining the ultimate goal. The iterative latent denoising in MMDiT accounts for \textbf{118.4s} of the total end-to-end latency, requiring 480 NFEs, albeit some on lower spatial resolutions. To further accelerate generation, we explore diffusion step-distillation techniques in this subsection, which aim to reduce the NFE requirement of the Denoising Scheduler. To this end, we begin by detailing the training objective of \textit{Pyramidal} Flow-Matching~\cite{jin2024pyramidal}, followed by an explanation of how we adapt four different Flow-Matching step-distillation techniques from the literature to this pyramidal setting. The selected techniques include DMD~\cite{yin2024one}, Direct Progressive Distillation~\cite{salimans2022progressive}, and Adversarial Distillation~\cite{sauer2024adversarial} as discrete step-distillation methods, and the recent Mean-Flows~\cite{geng2025mean} as a continuous consistency-based approach. While Mean-Flows was originally proposed for training models from scratch, we adapt it to operate in a distillation setting by applying it to an already trained model.

It is fascinating to observe how the training objectives for diffusion models have evolved---from early \textbf{Score Matching} approaches~\cite{hyvarinen2005estimation, song2019generative}, which required on the order of 1{,}000 denoising steps for generation, to the current state-of-the-art \textbf{Flow-Matching} methods~\cite{lipman2022flow, heitz2023iterative, liu2022flow}, which enable high-quality synthesis in as few as 50 steps. Along this trajectory, numerous influential works~\cite{ho2020denoising, song2020denoising, song2021score, karras2022elucidating} have contributed to making diffusion training both simpler and more scalable. Modern video diffusion models adopt a remarkably straightforward yet highly scalable training algorithm, capable of handling datasets with upwards of $\sim$100M videos. Starting from video latents $\vz \sim p_\text{data}(\vz)$ extracted via a fixed codec-latent-VAE, we construct noisy samples as
\[
\Tilde{\vz}_\sigma = (1 - \sigma)\vz + \sigma\epsilon,
\]
where $\sigma \sim \mathbb{U}(0, 1)$ and  denotes the noise level (0 = clean, 1 = fully noisy) and $\epsilon \sim \mathcal{N}(0, \mathbb{I})$ is Gaussian noise. This defines a continuous probability flow over $\Tilde{\vz}_\sigma$, whose instantaneous velocity is given by
\[
v(\Tilde{\vz}_\sigma) = \frac{d\Tilde{\vz}_\sigma}{d\sigma} = \epsilon - \vz.
\]
Interestingly, this velocity does not depend on $\sigma$ due to the linearity of the flow, in contrast to earlier formulations such as DDPM~\cite{ho2020denoising}. The MMDiT denoiser $\mathcal{D}$ is trained to predict these velocities, conditioned on the noise level $\sigma$, via the Flow Matching objective:
\[
\mathcal{L}_\text{FM} = \mathbb{E}_{\sigma,\vz,\epsilon}\big[\|\mathcal{D}(\Tilde{\vz}_\sigma, \sigma) - v(\Tilde{\vz}_\sigma)\|_2^2\big].
\]
Once trained, generation reduces to solving the ODE
\[
\vz = \epsilon - \int_{\sigma=1}^{0}\mathcal{D}(\Tilde{\vz}_\sigma, \sigma)\,d\sigma,
\]
typically using a first-order solver such as discrete Euler with 50 steps, though higher-order solvers are also applicable. In practice, $\mathcal{D}$ is further conditioned on text prompt embeddings $\hat{t}$ and CLIP embeddings $\hat{c}$\footnote{For clarity, we omitted explicit timestep conditioning in earlier sections.}.

Our base model \baseline~\cite{jin2024pyramidal} decomposes the probability flow into $S$ stages, where the $i^\text{th}$ stage ($i \in \{0,\dots,S-1\}$) operates at a resolution that is $2^{i}$-times smaller than the original. Let $\operatorname{Down}(\cdot, s)$ and $\operatorname{Up}(\cdot, s)$ denote downsampling and upsampling by a factor $s$, respectively. Each stage is parameterised by a pair of noise levels $(\sigma^i_{\mathrm{start}}, \sigma^i_{\mathrm{end}})$ with $1 > \sigma^i_{\mathrm{start}} > \sigma^i_{\mathrm{end}} > 0$, and operates on latents at resolution $\operatorname{Down}(\vz, 2^{i})$. The start and end distributions for stage $i$ are defined as
\begin{align}
\label{eq:stage_wise_noisy_start}
    \tilde{\vz}_{\sigma^i_{\mathrm{start}}}
    &:= (1 - \sigma^i_{\mathrm{start}})\,\operatorname{Up}\!\big(\operatorname{Down}(\vz, 2^{i+1}), 2\big)
       + \sigma^i_{\mathrm{start}}\,\epsilon, \\
    \tilde{\vz}_{\sigma^i_{\mathrm{end}}}
    &:= (1 - \sigma^i_{\mathrm{end}})\,\operatorname{Down}(\vz, 2^{i})
       + \sigma^i_{\mathrm{end}}\,\epsilon,
\label{eq:stage_wise_noisy_end}
\end{align}
where $\epsilon \sim \mathcal{N}(0,\mathbb{I})$.
By the universality of the Flow Matching objective, the stage-wise loss $\mathcal{L}^{i}_{\mathrm{FM}}$ is well defined to learn the  flow between the above start and end distributions at stage $i$. 
A different local noise-level $\sigma^i_\mathrm{local} \sim \mathbb{U}(0, 1)$ is used to learn the Flow-Matching model at the $i^\text{th}$ stage, and the global noise-level $\sigma^i$ relates to the local noise-level $\sigma^i_\mathrm{local}$ as $\sigma^i = \sigma^i_{\mathrm{start}}$. Thus, the overall Pyramidal Flow Matching objective is an aggregate over the stagewise objectives:
\[
\mathcal{L}_{\mathrm{pyr\mbox{-}FM}} := \sum_{i=0}^{S-1} \mathcal{L}^{i}_{\mathrm{FM}}.
\]
Intuitively, the Flow-Matching model at $i^\text{th}$ stage flows from a noisy and pixelated version of the latent video to less noisy and less pixelated version. Note that the model is not only learning the denoising objective, but also the super-resolution objective when matching the ground truth instantaneous flow. The most ingenious contribution from \baseline~is that the noise-levels of the different stages are not disjoint, but overlapping, which are obtained as:
\begin{align}
    \sigma^i_{\mathrm{end}} &= i.\frac{1}{S} \\
    \sigma^i_{\mathrm{start}} &= \frac{2\sigma^{i + 1}_{\mathrm{end}}}{1 + \sigma^{i + 1}_{\mathrm{end}}}
\end{align}
And thus, after training, the same MMDiT denoiser network $\mathcal{D}$ can be used to generate the samples by flowing through all the stages and jumping resolution across stages using the following equations: 
\begin{align}
    \tilde{\vz}_{\sigma^{i}_{\mathrm{start}}}
    &= \frac{1 + \sigma^{i}_{\mathrm{start}}}{2}\,
    \operatorname{Up}\!\big(\tilde{\vz}_{\sigma^{i+1}_{\mathrm{end}}},\,2\big)
    + \frac{\sqrt{3}\,\bigl(1 - \sigma^{i}_{\mathrm{start}}\bigr)}{2}\,\boldsymbol{\epsilon}' \\[3pt]
    \text{such that, } \epsilon' &\in \mathcal{N}(0, \Sigma') \text{ and } \Sigma'_\text{block} = [\text{Big } \gamma \text{ Matrix}]
\end{align}

\paragraph{Pyramidal Mean-Flows}
Mean-FLows~\cite{geng2025mean} proposed two changes to the learning algorithm of Flow-Matching models. First, they propose to model not the instantaneous velocity field $v(\Tilde{\vz}_\sigma)$ of the underlying probability-flow ODE, but instead the \textit{Mean} velocity field $v_\text{mean}(\Tilde{\vz}_\sigma, \beta, \sigma)$ which denotes the average velocity of the trajectory going from $\beta$ to $\sigma$ (such that, $\sigma > \beta$). A direct implication of which is that the denoiser network now needs to also condition on $\beta$ apart from $\sigma$, i.e. the mean-predicted velocity is now $\mathcal{D}(\Tilde{\vz}_\sigma, \beta, \sigma)$. Through a very interesting derivation, Mean-Flows shows that the learning objective for such a Mean-Flows model is given by:
\[
\mathcal{L}_\text{MF} = \mathbb{E}_{\sigma,\beta,\vz,\epsilon}\big[\|\mathcal{D}(\Tilde{\vz}_\sigma, \beta, \sigma) - v_\text{mean}(\Tilde{\vz}_\sigma, \beta, \sigma)\|_2^2\big].
\]
Where, the target ground-truth $v_\text{mean}$ is computed as: 
\[
v_\text{mean}(\Tilde{\vz}_\sigma, \beta, \sigma) = v(\Tilde{\vz}_\sigma, \sigma) - (\sigma - \beta)(v(\Tilde{\vz}_\sigma, \sigma)\partial_{\vz}\mathcal{D} + \partial_\sigma\mathcal{D})
\]
Where the latter term is computed as a \texttt{JVP} in code. In Our Pyramidal Mean-Flows version, we extend the Pyramidal Flow Matching loss for each of the $i^\text{th}$ stage such that the $v_\text{mean}$ is computed as, 
\[
    v_\text{mean-pyr} := v(\Tilde{\vz}_\sigma, \sigma) - (\sigma - \beta)(v(\Tilde{\vz}_\sigma, \sigma)\partial_{\vz}\mathcal{D} + (\sigma^i_\text{start} - \sigma^i_\text{end})\partial_\sigma\mathcal{D})
\]
Note the scaling of the last term, which accounts for the scaled version of the instantaneous velocities which are learned by the Pyramidal-Flow Matching model. Thus, with this correction to the $v_\text{mean-pyr}$, we can define the squared L2 loss per stage to obtain $\mathcal{L}^i_\text{MF-pyr}$, and then giving us the aggregate loss function $\mathcal{L}_\text{MF-pyr}$. In practice of course since we are finetuning the MMDiT $\mathcal{D}$ from a pretrained Flow-Matching model, we first only finetune it with the second starting point conditioning $\beta$ while only training for Flow-Matching objective, i.e. setting $\beta=\sigma$, and later training the Mean-Flows objective. Also as a key detail, the training needs to have only 25\% of the batch-samples that are Mean-Flows (i.e. $\beta \neq \sigma$), while the rest are still Flow-Matching samples so that the training stabilises. From our experiments, we found that fine-tuning the model in such a way very soon leads to a loss that is overpowered by Flow-Matching rather than Mean-Flows, while increasing more Mean-Flows batch samples leads to unstable training. Further exploring the cause of this instability constitues an interesting direction for future work.

\paragraph{Pyramidal DMD}
To unlock the few-step inference regime of our model, we adapted the step distillation pipeline of DMD~\cite{yin2024one} to PyramidalFlow's stage-wise inference.
At $i^\text{th}$ stage for the input 
$\tilde{\vz}_\sigma =  (1 - \sigma^i_{\mathrm{local}}) \tilde{\vz}_{\sigma^i_{\mathrm{end}}} + \sigma^i_{\mathrm{local}} \tilde{\vz}_{\sigma^i_{\mathrm{start}}} $ the student network $\mathcal{D}_\theta (\Tilde{\vz}_\sigma, \sigma)$ aims to predict the clean latent, parametrized as a single-step Euler solver $\Tilde{\vz}_\theta := \tilde{\vz}_\sigma - \frac{\sigma}{\sigma^i_{\mathrm{start}} - \sigma^i_{\mathrm{end}}} \mathcal{D}_\theta (\Tilde{\vz}_\sigma, \sigma).$

Such scaling of network output has been chosen since student's weights are initialized with the pretrained teacher $\mathcal{D}$. 
The teacher had been trained to approximate the flow defined as a derivative w.r.t. $\sigma^i_{\mathrm{local}}.$
Since $\sigma = (1 - \sigma^i_{\mathrm{local}}) \sigma^i_{\mathrm{end}} + \sigma^i_{\mathrm{local}} \sigma^i_{\mathrm{start}}, $ the derivative w.r.t the global noise level $\sigma$ should be scaled by $\frac{d\sigma^i_{\mathrm{local}}}{d\sigma} = \frac{1}{\sigma^i_{\mathrm{start}} - \sigma^i_{\mathrm{end}}}.$

The so-called \emph{fake model} $\mathcal{D}_\varphi $ is trained with pyramidal Flow Matching objective $\mathcal{L}_{\mathrm{pyr\mbox{-}FM}}$  defined above but on the distribution of student-predicted clean latents instead of ground-true video latents.
Having the fake model, the student network is updated with DMD loss defined through its gradient
$\nabla_\theta L_{\text{DMD}}^i \propto \left(  \mathcal{D} (\Tilde{\vz}_\tau, \tau) - \mathcal{D}_\varphi (\Tilde{\vz}_\tau, \tau)  \right) \cdot \nabla_\theta \Tilde{\vz}_\theta.$
The input of teacher and fake model $\Tilde{\vz}_\tau$ is defined as a stage-wise noisy version of student-predicted clean latent, similar to~\cref{eq:stage_wise_noisy_start,eq:stage_wise_noisy_end},
\begin{align}
    \tilde{y}_{\sigma^i_{\mathrm{start}}}
    &:= (1 - \sigma^i_{\mathrm{start}})\,\operatorname{Up}\!\big(\operatorname{Down}(\Tilde{\vz}_\theta , 2), 2\big)
       + \sigma^i_{\mathrm{start}}\,\varepsilon, \\
    \tilde{y}_{\sigma^i_{\mathrm{end}}}
    &:= (1 - \sigma^i_{\mathrm{end}})\,\Tilde{\vz}_\theta 
       + \sigma^i_{\mathrm{end}}\,\varepsilon, \\
    \tilde{\vz}_\tau &:=  (1 - \tau^i_{\mathrm{local}}) \tilde{y}_{\sigma^i_{\mathrm{end}}} + \tau^i_{\mathrm{local}} \tilde{y}_{\sigma^i_{\mathrm{start}}},
\end{align}
where $\varepsilon \sim \mathcal{N}(0,\mathbb{I})$ and $\tau = (1 - \tau^i_{\mathrm{local}}) \sigma^i_{\mathrm{end}} + \tau^i_{\mathrm{local}} \sigma^i_{\mathrm{start}}$.

We follow \cite{yin2024one} and define the sample-specific weight of DMD loss as $\left\lVert \mathcal{D} (\Tilde{\vz}_\tau, \tau) - \left(\tilde{y}_{\sigma^i_{\mathrm{start}}} - \tilde{y}_{\sigma^i_{\mathrm{end}}} \right) \right\rVert_1^{-1}.$ Therefore, the sample gets higher weight, if teacher model is capable to estimate its conditional flow with a smaller error.
In addition we found the supervised Cauchy loss $L_{\text{teacher}} = \log\left(1 + \left\lVert \Tilde{\vz}_\theta - \operatorname{Down}(\vz, 2^{i})\right\rVert_2^2 \right)$ useful for visual quality and used it with weight 0.5.
During training, we update the student and fake model in alternate manner: one update of  $\theta$ per two updates of $\varphi$.
For student's updates we limit the set of local noise levels $\tau^i_{\mathrm{local}}$ to four evenly selected values, and for fake model it is sampled from $\mathbb{U}(0, 1).$
To obtain teacher's prediction we employ classifier-free guidance with the same hyperparameters as those recommended for the sampling from that model.
The whole training required 5000 iterations on 16 H100 GPUs.

In our early experiments on step distillation we found that training process was unstable if the original pretrained model was used as a teacher for the student at our target spatial resolution. 
For that reason, we employed the lower-resolution checkpoint both for the teacher model and as an initialization for the student and fake model.
At inference time, generating images with the student model is done in the same way as with the teacher: we use Euler sampler but only with a few steps per each stage.

\paragraph{Pyramidal Progressive Distillation}
We apply the Progressive Distillation~\cite{salimans2022progressive} to Pyramidal Flow-Matching. Firstly, we found that stage-wise 2$\times$ distillation is not necessary for a Pyramidal Flow-Matching model, which already can do the generation with 20 steps for a single resolution, thereby doing denoising of a single frame in 60 steps (given 3 pyramidal resolutions). Thus, for per stage we setup two networks: a student network $\mathcal{D}_\theta$ (to be distilled) and the teacher network $\mathcal{D}$ for supervision. Then, we obtain synthetic generated videos for the $\sim$350K prompts that we had curated to form our dataset so that we never leave the support of the probability distribution that is learned by the teacher network $\mathcal{D}$.

\begin{table}[!t]
\centering
\caption{\textbf{Quantitative Evaluation of Step Distillation.} We compare various step-distillation methods adapted to Pyramidal Flow-Matching for efficient video generation across two different inference schedules; namely 4-4-4 and 1-1-1. Except for the Pyramid-Flow Native, which generates first frame with 20-20-20 steps and rest of the frames with 10-10-10, all generate the \texttt{[7x40x64]} latent video with the same number of steps for first and rest of the frames. The metrics for comparison include \textbf{VBench} scores—\textbf{Total} (Tot.$\uparrow$), Visual \textbf{Quality} (Qual.$\uparrow$), \textbf{Semantic} Score (Sem.$\uparrow$), and \textbf{Dynamic Degree} (D.D. $\uparrow$).}
\label{tab:step_distil_quant_main}
\resizebox{\textwidth}{!}{%
\begin{tabular}{|l|c|c|c|c|c|c|c|c|}
\hline
\textbf{Step} & \textbf{Inference} & \textbf{Distillation} & \textbf{Total} &\textbf{ Qualcomm} & \multicolumn{4}{c|}{\textbf{VBench Score}} \\
\cline{6-9}
 \textbf{Distillation} & \textbf{Schedule} & \textbf{Schedule} & \textbf{MMDiT} & \textbf{Hexagon NPU} & \textbf{Tot.($\uparrow$)} & \textbf{Qual.($\uparrow$)} & \textbf{Sem.($\uparrow$)} & \textbf{D.D.($\uparrow$)} \\
\textbf{Method} &  &  & \textbf{NFEs ($\downarrow$)} & \textbf{Latency ($\downarrow$)} &  &  & &  \\

\hline
\cellcolor{tabbaseline}Pyramidal-Flow Native &\cellcolor{tabbaseline}(20)10$\times$3 &\cellcolor{tabbaseline}N/A &\cellcolor{tabbaseline}480 &\cellcolor{tabbaseline}118.40s & \cellcolor{tabbaseline} 80.31 & \cellcolor{tabbaseline} 83.68 & \cellcolor{tabbaseline} 66.81 & \cellcolor{tabbaseline} 64.72 \\
\cellcolor{tabbaseline}(CFG present)&\cellcolor{tabbaseline}4-4-4 &\cellcolor{tabbaseline}N/A &\cellcolor{tabbaseline}168 &\cellcolor{tabbaseline}41.44s & \cellcolor{tabbaseline} 75.82 & \cellcolor{tabbaseline} 79.90 & \cellcolor{tabbaseline} 59.49 & \cellcolor{tabbaseline} 49.17 \\
\cellcolor{tabbaseline} &\cellcolor{tabbaseline}1-1-1 &\cellcolor{tabbaseline}N/A &\cellcolor{tabbaseline}42 &\cellcolor{tabbaseline}10.36s & \cellcolor{tabbaseline} 59.62 & \cellcolor{tabbaseline} 67.90 & \cellcolor{tabbaseline} 26.50 & \cellcolor{tabbaseline} 6.39 \\
\hline

Pyramidal Mean-Flows  & 4-4-4 & N/A & 168 & 41.44s & 76.25 & 80.60 & 58.87 & 49.17 \\
(CFG present) & 1-1-1 & N/A & 42  & 10.36s & 63.44 & 70.89 & 33.62 & 20.00 \\
\hline

Pyramidal DMD & 4-4-4 & 1-1-1 & 84 & 20.72s  & \cellcolor{tabfirst} 80.37  & \cellcolor{tabfirst} 85.21  & \cellcolor{tabsecond} 61.01  & \cellcolor{tabsecond} 86.11 \\
    & 1-1-1 & 1-1-1 & 21  & 5.18s  & \cellcolor{tabsecond} 76.48  & \cellcolor{tabsecond} 80.79  & \cellcolor{tabsecond} 59.24  & \cellcolor{tabthird} 48.89 \\
\hline

Pyramidal Progressive & 4-4-4 & 4-4-4 & 84 & 20.72s & \cellcolor{tabthird} 78.22 & \cellcolor{tabthird} 82.41 & \cellcolor{tabfirst} 61.46 & \cellcolor{tabthird} 52.50 \\
                                & 1-1-1 & 1-1-1 & 21 & 5.18s & \cellcolor{tabthird} 76.17  & \cellcolor{tabthird} 80.46 & \cellcolor{tabthird} 59.02 & \cellcolor{tabsecond} 62.22 \\
\hline

Pyramidal Adversarial  & 4-4-4 & 4-4-4 & 84 & 20.72s & \cellcolor{tabsecond} 78.51 & \cellcolor{tabsecond} 83.19 & \cellcolor{tabthird} 59.77 & \cellcolor{tabfirst} 87.78 \\
                               & 1-1-1 & 1-1-1 & 21 & 5.18s & \cellcolor{tabfirst} 77.47 & \cellcolor{tabfirst} 81.74 & \cellcolor{tabfirst} 60.39 & \cellcolor{tabfirst} 64.17 \\
\hline

\end{tabular}
}
\end{table}

We quantise the steps of the Teacher model to 16 uniformly sampled $\sigma^i \in \{\sigma^i_\text{start}, ... \text{14 steps} ... ,\sigma^i_\text{end}\}$ values (Euler solver) for the $i^\text{th}$ stage. The Student then either learns a single step (for 1-1-1 configuration) or 4 steps (for 4-4-4 configuration) uniformly subsampled from the set of the teacher steps. We detail the loss objective for the 4-4-4 configuration below, but note that the 1-1-1 follows directly from it, or any other distillation configuration as long as the number of teacher steps are are perfectly divisible by the number of student steps that are to be distilled.

Given a sampled $\sigma^{is}_\text{teach}$, we take the next $j$ steps (in this case 4 steps) to fix the last sigma from the teacher inference as $\sigma^{is+j}_\text{teach}$. The ground truth end-point from the teacher inference trajectory given the starting point $\Tilde{\vz}_{\sigma^{is}_\text{teach}}$ is computed by running \texttt{no\_grad} mode Euler inference of the Teacher:
\[
    \Tilde{\vz}_{\sigma^{is+j}_\text{teach}} = \Tilde{\vz}_{\sigma^{is}_\text{teach}} + \sum_s^{s+j} (\sigma^{is+1} - \sigma^{is})\mathcal{D}(\Tilde{\vz}_{\sigma^{is}_\text{teach}})
\]

Then, the student model's prediction to match the $\Tilde{\vz}_{\sigma^{is+j}_\text{teach}}$ in one step is computed again as an Euler step: 
\[
    \Tilde{\vz}_{\sigma^{is+j}_\text{stud}} = \Tilde{\vz}_{\sigma^{is}_\text{stud}} + (\sigma^{is+j} - \sigma^{is})\mathcal{D}_\theta(\Tilde{\vz}_{\sigma^{is}_\text{stud}})
\]

Thus, having computed the noisy simulations from the Teacher as well as the student's one-step noisy predictions, the loss function for this distillation is defined as
\[
    \mathcal{L}_\text{pyr-prog} := \mathbb{E}_{\sigma,\vz,\epsilon}\big[\| \Tilde{\vz}_{\sigma^{is+j}_\text{stud}} - \Tilde{\vz}_{\sigma^{is+j}_\text{teach}} \|_2^2\big]
\]
Intuitively, we basically run the teacher model $\mathcal{D}$ for four Euler steps to get the teacher's prediction, without gradients and then train the student model to match this output. Once distilled, the student model can then be used to run the inference in the distilled number of steps. While in all this, as explained mathematically, we ensure that the velocities are still being scaled per resolution (stage) correctly and that we use the local noise-levels in the teacher and student simulations. 

\begin{figure}[!t]
  \centering
  \includegraphics[width=0.8\textwidth]{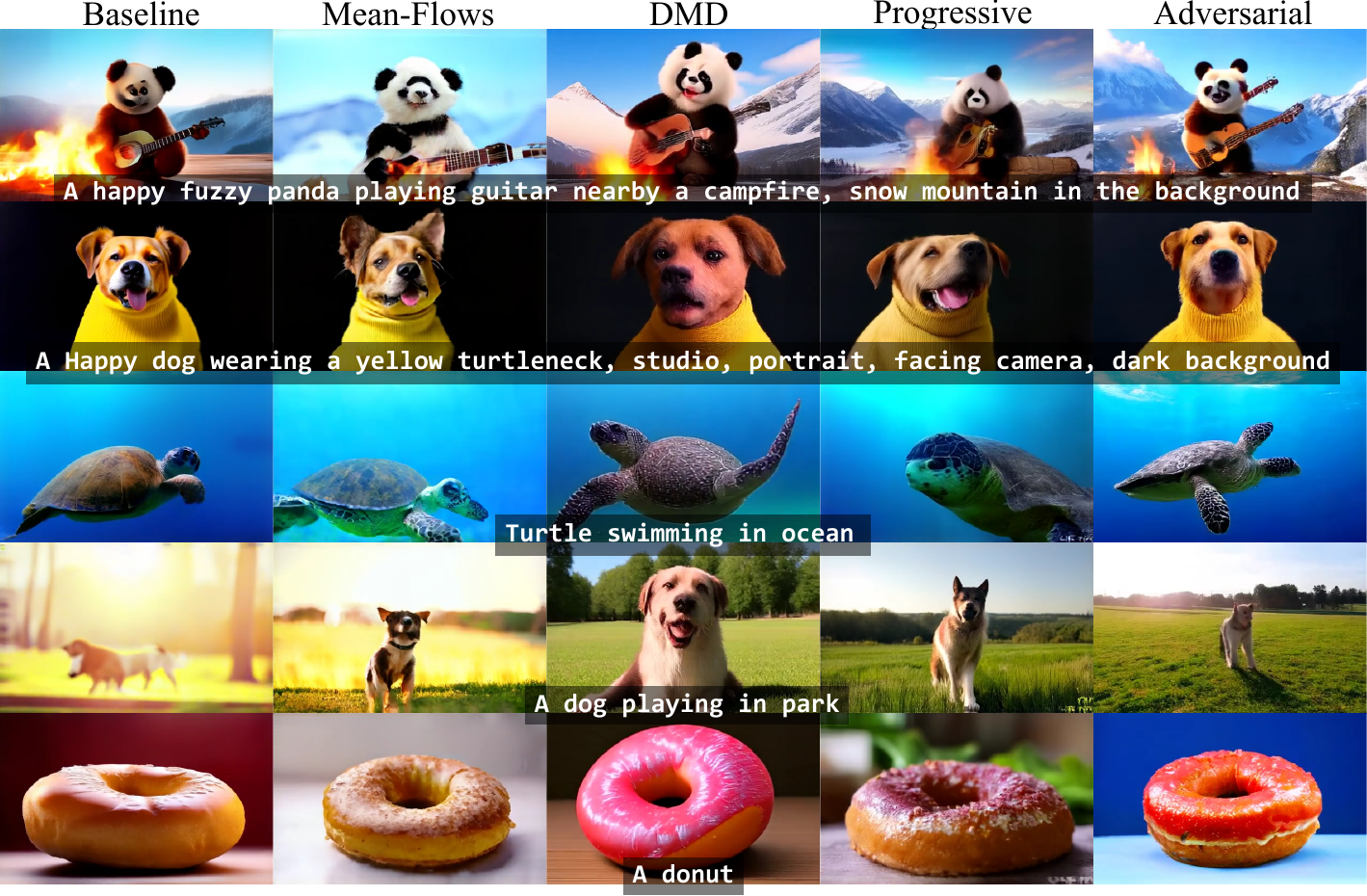}
  \captionof{figure}{\textbf{Qualitative Evaluation of Step Distillation.} We visualise randomly selected frames from the generated \texttt{[49×320×512]} videos, across the block-pruned model distilled using different step-distillation techniques, namely Mean-Flows, DMD, Progressive, and Adversarial step distillation.}
  \label{fig:step_distil_qual_main}
\end{figure}

\paragraph{Pyramidal Adversarial Step Distillation}
Finally in the Pyramidal Adversarial Step Distillation approach, we follow the same setup as the progressive one, but also add a patchwise GAN loss on top of the squared L2 loss of the Progressive distillation approach.
\[
    \mathcal{L}_\text{pyr-adv} := w_\text{recon}\mathcal{L}_\text{pyr-prog} + w_\text{adv}\mathcal{L}_\text{GAN}(\Tilde{\vz}_{\sigma^{is+j}_\text{stud}}, \Tilde{\vz}_{\sigma^{is+j}_\text{teach}})
\]
We use the Hinge-GAN loss for the $\mathcal{L}_\text{GAN}$ which speeds up the distillation process by not only focussing on the pixelwise distance, but also matching the distributions of the noisy tokens adversarially. 

As is standard practice with Adversarial step distillation, we use the Features extracted from the Teacher network $\mathcal{D}$ passed to a 4layer MLP as the Discriminator architecture for the GAN loss. The teacher network always remains frozen and only the MLP is trainable. We empirically found the values of $w_\text{recon} = 10.0$ and $w_\text{GAN} = 1.0$ to converge well. 

\paragraph{Results Analysis.} Table~\ref{tab:step_distil_quant_main} and Figure~\ref{fig:step_distil_qual_main} summarise our quantitative and qualitative results respectively. 
Since the pyramidal setting of the base model operates on 3 different resolutions at the time of generation, we specifically target two different configurations, namely 4-4-4 and 1-1-1, where the denoiser spends 4 steps and 1 step on the three denoising resolutions respectively. Except for Pyramidal Mean-Flows, all the other three adaptations provide significant VBench gains compared to the base non-distilled model's performance, especially in the lower step regime of 1-1-1. Since Pyramidal DMD yields the best VBench score of \textbf{80.37} for the 4-4-4 setting, we chose to use this step-distilled MMDiT denoiser $\mathcal{D}$, with a denoising latency of \textbf{20.72s} for the \name pipeline (see appendix fig.~\ref{fig:final_full_pipeline_illustration}). But, in the native T2V setting, we noticed significant amount degradation in the visual quality of the generated videos. There were two main artifacts: one in the form of semantic misalignment as well as the typical colour saturation artifacts that are very common in DMD based step-distillation methods. In the next section, we describe not only how we get rid of these artifacts, but also detail the nuances of integrating all our optimisations together into a single coherent end-to-end mobile video generation pipeline.

\section{End-to-End Integration}

By applying step distillation, we reduced the end-to-end video generation latency to \textbf{20.72}, bringing our system close to the threshold of interactive video generation at \textbf{2fps}. While the model maintains a strong VBench score of \textbf{80.37}, we observe noticeable visual-semantic degradation. As illustrated in Figure~\ref{fig:step_distil_qual_main}, our final \textit{Pyramidal}-DMD approach introduces colour saturation artifacts and semantic inconsistencies in the first frame. Nevertheless, the generated motion remains smooth and stable, suggesting that this issue is not fully captured by the VBench~\cite{huang2024vbench} evaluation suite. We hypothesize that these semantic artifacts can be mitigated by initializing the video with a high-quality first frame generated by a separate text-to-image model. This strategy would preserve the integrity of the initial frame while leveraging our pipeline to apply coherent and descriptive motion to subsequent frames.

\begin{figure}[!t]
  \centering
  \includegraphics[width=0.8\textwidth]{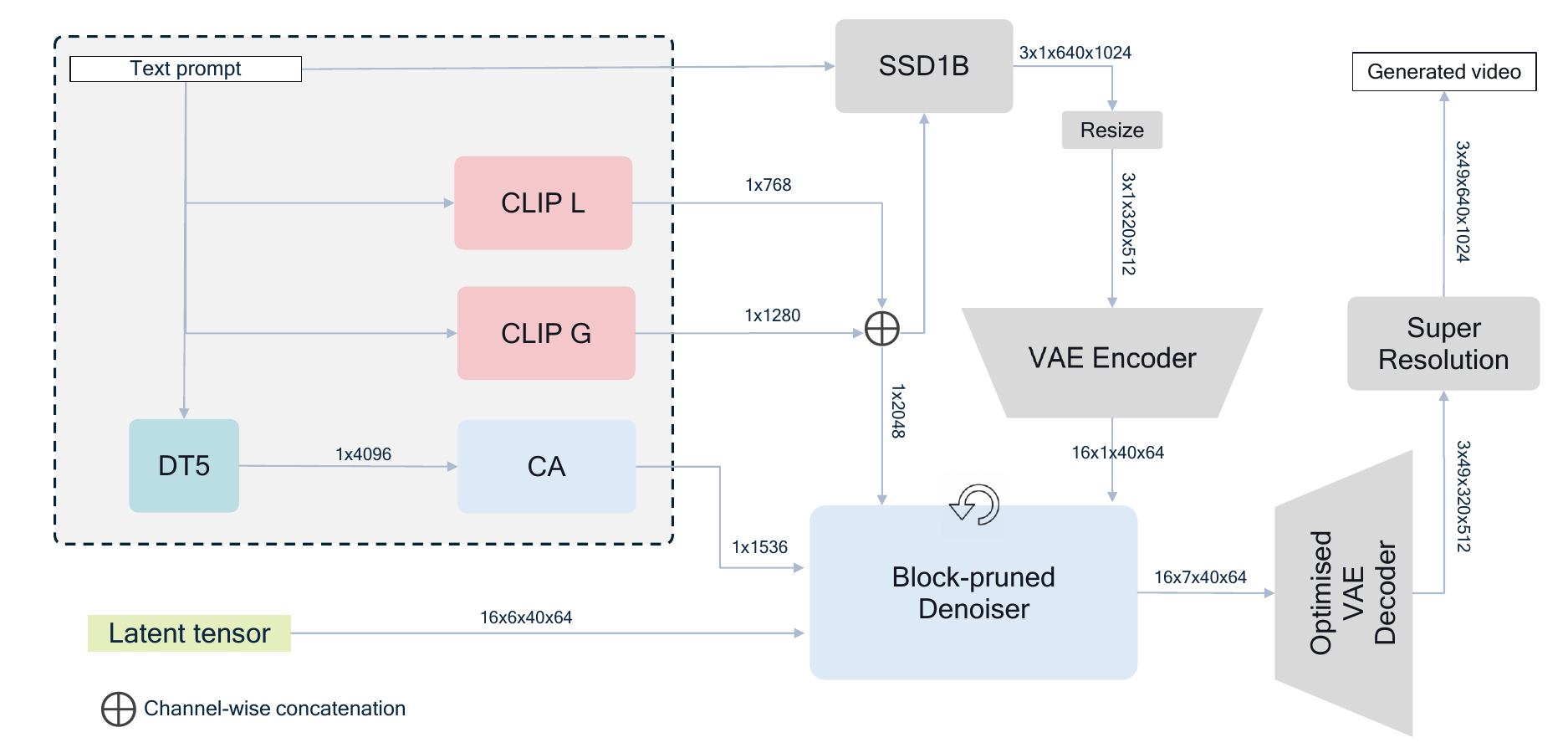}
  \captionof{figure}{\textbf{Overview of the \name~E2E pipeline}. As opposed to the base pipeline detailed in figure \ref{fig:base_pipeline_illustration}, the final E2E pipeline of \textbf{\name} integrates all four of our proposed optimisations namely, Distilled small Text-Encoder \ref{subsec:text_enc_distil}, Asymmetric VAE Decoder \ref{subsec:asymm_dec_distil}, Block-Pruned MMDiT \ref{subsec:block_pruning}, and Step-Distilled scheduler \ref{subsec:step_distil} (not shown here). For boosting the visual fidelity of the generations, we also include SSD1B \cite{gupta2024progressive} for generating the first image, and QuickSRNet \cite{berger2023quicksrnet} for $2\!\times\!$ super-resolution.}
  \label{fig:final_full_pipeline_illustration}
\end{figure}

\begin{table}[!t]
\centering
\caption{\textbf{\name~On-device Measurements}. We report on-device measurements for running our proposed \name~system on the laptop SoC \laptopsoc~and the mobile SoC \mobilesoc; both powered by the \platform. Since the VAE Encoder is only run once for the first frame it is unoptimised and we also report the peak RAM usage of each component for the latop SoC.}
\label{tab:neodragon_on_device}
\resizebox{\textwidth}{!}{%
\begin{tabular}{|l|c|c|c|c|c|c|c|c|c|c|c|}
\hline
\textbf{SoC} & \textbf{CLIP} & \textbf{CLIP} & \textbf{DistilT5} & \textbf{VAE} & \textbf{VAE} & \multicolumn{2}{c|}{\textbf{SSD1B}} & \multicolumn{3}{c|}{\textbf{MMDiT+CA}} & \textbf{QuickSRNet} \\ 
\cline{7-11}
\textbf{/ Measurement} & \textbf{L} & \textbf{G} & & \textbf{Enc.} & \textbf{Dec.} & \textbf{UNet} & \textbf{Dec.} & \textbf{[\texttt{7$\times$10$\times$16}]} & \textbf{[\texttt{7$\times$20$\times$32}]} & \textbf{[\texttt{7$\times$40$\times$64}]} & \\ 
 \hline
\textbf{\laptopsoc~/Lat. ms} & 5.9 & 43.6 & 3.0 & 941.7 & 143.0 & 151.5 & 378.6 & 54.9 & 101.4 & 590.2 & 4.9 \\ 
 \hline
\textbf{\laptopsoc~/Mem. GB} & 0.49 & 2.64 & 0.03 & 0.68 & 0.21 & 2.57 & 0.17 & 3.13 & 3.15 & 3.25 & 0.01 \\ 
 \hline
\textbf{\mobilesoc~/Lat. ms} & 14.0 & 76.5 & 3.5 & 1206.5 & 248.9 & 234.6 & 580.0 & 104.7 & 218.3 & 938.3 & 6.5 \\ 
 \hline
\end{tabular}
}
\end{table}

\begin{figure}
  \centering
  \includegraphics[width=\textwidth]{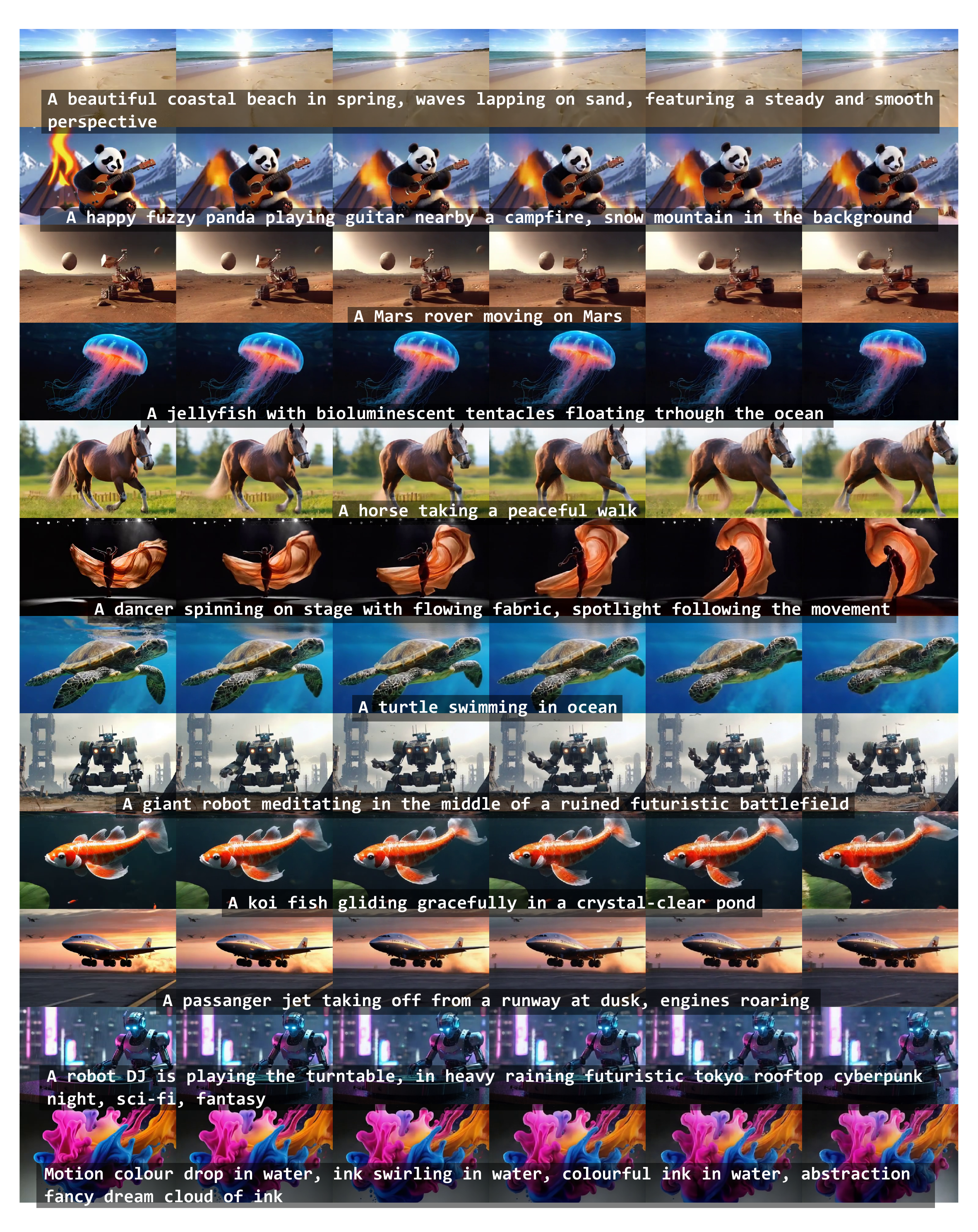}
  \captionof{figure}{\textbf{\name} produces high-fidelity videos with strong semantic alignment to input prompts. Shown here is a sampler of generations spanning complex motions, diverse scene compositions, and both realistic and imaginative content.}
  \label{fig:final_e2e_qualitative}
\end{figure}

To validate this hypothesis, we initialize the first frame using SSD-1B~\cite{gupta2024progressive} in four steps and then generate the remaining latent frames using our optimized pipeline. As shown in Figure~\ref{fig:final_e2e_qualitative}, this approach produces high-fidelity videos with smooth motion. Interestingly, when applying the 1-1-1 step-distilled \textit{Pyramidal}-DMD model with the first frame from SSD-1B, we observe strong performance while further reducing end-to-end latency. This configuration introduces notable changes to the latency profile: SSD-1B generates the first frame in four steps, requiring \textbf{0.82s}, including the time for CLIP embedding extraction via its text encoder. We integrate this text encoder into a unified SSD-1B module within our final pipeline (see Fig.~\ref{fig:final_full_pipeline_illustration}). To obtain the first-frame latents, we additionally run our fixed VAE encoder, which takes approximately \textbf{0.94s} . Accounting for these steps and using the 1-1-1 distilled model for subsequent frame generation, the proposed \name~system achieves an end-to-end latency of \textbf{6.6s}. To further enhance visual quality, we apply 2$\times$ supersampling using QuickSRNet~\cite{berger2023quicksrnet}, which adds only \textbf{5ms} to the pipeline, resulting in a final E2E latency of approximately \textbf{6.7s} on the \laptopsoc. Table \ref{tab:neodragon_on_device} provides details of the on-device measurements done for running our \name~system on both a Laptop SoC and a Mobile SoC; both containing the \platform. Table \ref{tab:sota_comparison_table} summarizes the leaderboard of open video generation models, where our system ranks at the top among on-device solutions, achieving the highest VBench score of \textbf{81.61}.

\begin{table}[ht]
\centering
\small
\setlength{\tabcolsep}{2pt} 
\renewcommand{\arraystretch}{1.05} 
\caption{\textbf{Comparison with state-of-the-art video generation models}. All \textbf{VBench} scores from the compared methods are extracted from their reported numbers, except for `Wan2.1*', and 'Pyramidal-Flow$^{\antichrist}$' which are our reproduction of the scores using the same evaluation pipelines and parameters for the genrated video resolution of \texttt{[49$\times$320$\times$512]} as we have used for our optimisations.}
\label{tab:sota_comparison_table}
\resizebox{\textwidth}{!}{
\begin{tabular}{|c|l|c|c|c|c|c|c|c|c|c|}
\hline
\textbf{Platform} & \textbf{Model} & \textbf{Tot.($\uparrow$) }& \textbf{Qual.($\uparrow$)} & \textbf{Sem.($\uparrow$)} & \textbf{Flick.($\uparrow$)} & \textbf{Aes.($\uparrow$)} & \textbf{Imag.($\uparrow$)} & \textbf{Obj. ($\uparrow$)} & \textbf{Scene($\uparrow$)} & \textbf{Cons.($\uparrow$)} \\
\hline
&Pyramidal-Flow & 81.72 & 84.74 & 69.62 & 99.49 & 63.26 & 65.01 & 86.67 & 43.20 & 26.23 \\
&Wan2.1 1.3B  & 83.31 & 85.23 & 75.65 & 99.55 & 65.46 & 67.01 & 88.81 & 41.96 & 25.50 \\
&Wan2.1 1.3B* & 82.47 & 83.33 & 79.01 & 99.35 & 65.05 & 64.81 & 89.87 & 54.20 & 26.95 \\
&Open-Sora V1.2 & 79.76 & 81.35 & 73.39 & 99.53 & 56.85 & 63.34 & 82.22 & 42.44 & 26.85 \\
&CogVideoX1.5 5B & 82.01 & 82.72 & 79.17 & 98.53 & 62.07 & 65.34 & 83.42 & 53.28 & 27.42 \\
\textbf{Server} & CogVideoX 5B & 81.91 & 83.05 & 77.33 & 78.97 & 61.88 & 63.33 & 85.07 & 51.96 & 27.65 \\
&CogVideoX 2B & 81.55 & 82.48 & 77.81 & 98.85 & 61.07 & 62.37 & 86.48 & 50.04 & 27.33 \\
&Mochi-1  & 80.13 & 82.64 & 70.08 & 99.40 & 56.94 & 60.64 & 86.51 & 36.99 & 25.15 \\
&LTX-Video & 80.00 & 82.30 & 70.79 & 99.34 & 59.81 & 60.28 & 83.45 & 51.07 & 25.19 \\
&Efficient VDiT & 76.14 & - & - & 99.49 & 57.21 & - & 60.33 & - & - \\
& Attention Surgery - 15$\times$R2 & 83.21 & 85.19 & 75.25 & 99.48 & 65.58 & 67.48 & 88.53 & 42.79 & 25.21 \\

\hline
&\cellcolor{tabbaseline}Pyramidal-Flow$^{\antichrist}$ &\cellcolor{tabbaseline}80.31 &\cellcolor{tabbaseline}83.68 &\cellcolor{tabbaseline}66.81 &\cellcolor{tabbaseline}99.46 &\cellcolor{tabbaseline}60.40 &\cellcolor{tabbaseline}63.90 &\cellcolor{tabbaseline}84.79 &\cellcolor{tabbaseline}45.74 &\cellcolor{tabbaseline}26.31 \\
&Snap Mobile Video DiT &\cellcolor{tabsecond}81.45 & 83.12 &\cellcolor{tabfirst}74.76 & 98.11 &\cellcolor{tabthird}64.16 & 63.41 & 92.26 &\cellcolor{tabthird}51.06 & 25.51 \\
&Hummingbird 16frame &\cellcolor{tabthird}81.35 & \cellcolor{tabfirst}83.73 &\cellcolor{tabthird}71.84 & 95.24 &\cellcolor{tabfirst}68.04 &\cellcolor{tabfirst}71.04 &\cellcolor{tabfirst}96.36 &\cellcolor{tabsecond}52.91 &\cellcolor{tabfirst}28.09 \\
\textbf{Mobile} &Hummingbird 26frame & 80.31 & 83.11 & 69.10 & 97.64 &\cellcolor{tabsecond}67.82 &\cellcolor{tabsecond}69.94 &\cellcolor{tabsecond}94.86 & 49.49 &\cellcolor{tabsecond}27.65 \\
&SnapGenV & 81.14 & 83.47 &\cellcolor{tabthird}71.84 &\cellcolor{tabsecond}99.37 & 62.19 & - & 92.22 & - &\cellcolor{tabthird}27.42 \\
&\textbf{(Ours)} \textbf{\name} E2E &\cellcolor{tabfirst}81.61 & \cellcolor{tabsecond}83.68 &\cellcolor{tabsecond}73.36 &\cellcolor{tabthird}99.27 & 60.71 & 59.78 &\cellcolor{tabthird}92.37 &\cellcolor{tabfirst}56.56 &\cellcolor{tabfirst}28.09 \\
&\textbf{(Ours)} T2V Multi-Step & 80.21 &\cellcolor{tabthird}83.54 & 66.90 &\cellcolor{tabfirst}99.40 & 59.86 &\cellcolor{tabthird}63.56 & 83.16 & 42.54 & 26.58 \\
\hline
\end{tabular}
}
\end{table}

Thus, we began with the \baseline~\cite{jin2024pyramidal} generation pipeline, as illustrated in Figure~\ref{fig:base_pipeline_illustration}, and through a series of systematic optimisations, transformed it into the end-to-end pipeline shown in Figure~\ref{fig:final_full_pipeline_illustration}. Every component has been refined, replaced, or reimagined—yet the essence of the original design persists. In this sense, our work echoes the paradox of the Ship of Theseus: when every part of a system has changed, does it remain the same entity? We embrace this philosophical question and christen this new incarnation of the pipeline as \name, a system that carries forward the spirit of its predecessor while embodying an entirely renewed form. This transformation is not merely an engineering exercise; it reflects a deeper truth about progress in AI research: identity is not static but emergent, shaped by continuous adaptation and reinvention. 

\subsection{Model Compilation}
\label{subsec:pipe_odp}

Deploying the model on a fixed-point NPU involves the following model compilation steps:
 
\paragraph{Porting multi-resolution DiTs.} To compile static graphs, we need to port 3 DiT graphs-low, mid, and high- corresponding to the latent resolution of each stage. As illustrated in figure 2, the PyTorch model takes past history which dynamically grows with more latent frames being included. Thus, input paddings are needed for each DiT graph. Specifically, we expand the last frame graph and pad zeros when running the inference for earlier frames. Doing such requires changes to the attention mask and positional embedding implementation such that the zero paddings are not contributing to the next frame generation
 
\paragraph{Precomputation of DiT inputs.} the input merge function of DiT's forward pass computes constant information for each stage. These include input shapes and trainable token lengths specific to the current stage. Besides, as we run a fixed amount of timesteps, we also have timestep embeddings precomputed as inputs to the DiT graph
 
\paragraph{Reduction of 6D tensors.} The DiT graph contains temporal and spatial dimensions, which makes the RoPE layer computation 6D tensor mul/add. Unlike torch dynamic graphs, high dimensional inputs means more complicated tiling which usually ends up penalizing performance. In order to simplify this, we reduce along the sin/cos and broadcast dimensions. This enables an extreme performance boost. It reduces DiT-high compilation time from near a day to <2h. Gaining latency performance from seconds to sub-second
 
\paragraph{Optimizing causal mask value.} By default, the causal mask value in T5 and DiT is set to an extremely large number, which can cause problems during device deployment. To mitigate this, we adjust the mask value to a more suitable number—large enough to prevent the model from attending to masked tokens, but small enough to avoid overflow.
 
\paragraph{Rescaling activations in T5.} T5 includes \texttt{res\_add} and \texttt{ff\_add} operations with values exceeding the FP16 numerical range. To ensure numerical stability and maintain functional equivalence, we apply a scaling factor to these residual connections, effectively transforming them without altering the model behavior.

\subsection{Pipeline Quantization}
\label{subsec:pipe_quant}

\paragraph{Fixed point quantization.} The quantization scheme used for deploying the various modules and calibration results are described in table ~\ref{tab:quant_scheme}. We use \emph{AIMET - AI Model Efficiency Toolkit}~\cite{siddegowda2022neuralnetworkquantizationai} to perform Post Training Quantization (PTQ). W8A16 quantization scheme is used for all but QuickSRNet. We apply AIMET \emph{AdaRound}~\cite{pmlr-v119-nagel20a}, a weight rounding technique, to QuickSRNet which ends up adding 7+dB SQNR. In this table, SQNR is calculated between original FP models vs. the deployed models

It's worth pointing out that the complexity of our pipeline has an impact on end-to-end integration SQNR as the quantization loss, despite minimal for each model, compounds quickly. In DiT pipeline, for example, we have FP embeddings coming from text encoders and first frame generated by SSD1B pipeline. Then it goes through 3 stages of DiTs mutiplied by the number of timesteps run in each stage. After which it's decoded by VAE decoder and upsampled by QuickSRNet. This is where the step distillation mentioned in previous section comes to our assistance . Reduction to only 1 timestep on each stage doesn't just help reduce latency but also greatly mitigats the compounding quantization loss along the pipeline

\begin{table}[!t]
\centering
\caption{\textbf{\name~Quantization Scheme Used}. We report the quantization scheme used for each modules here.}
\label{tab:quant_scheme}
\resizebox{\textwidth}{!}{%
\begin{tabular}{|l|c|c|c|c|c|c|c|c|c|c|c|}
\hline
\textbf{Module}& \textbf{CLIP} & \textbf{CLIP} & \textbf{DistilT5} & \textbf{VAE} & \textbf{VAE} & \multicolumn{2}{c|}{\textbf{SSD1B}} & \multicolumn{3}{c|}{\textbf{MMDiT+CA}} & \textbf{QuickSRNet} \\ 
\cline{7-11}
& \textbf{L} & \textbf{G} & & \textbf{Enc.} & \textbf{Dec.} & \textbf{UNet} & \textbf{Dec.} & \textbf{[\texttt{7$\times$10$\times$16}]} & \textbf{[\texttt{7$\times$20$\times$32}]} & \textbf{[\texttt{7$\times$40$\times$64}]} & \\ 
 \hline
\textbf{Quant Scheme}& FP16& FP16& FP16& W8A16, PTQ & W8A16, PTQ & W8A16, PTQ & W8A16, PTQ & W8A16, PTQ & W8A16, PTQ & W8A16, PTQ & W8A16, AdaRound\\ 
 \hline
 \textbf{Calibration Data Size}& NA& NA& NA& 50& 50& 500& 500& 300& 300& 300& 500\\ 
 \hline
 \textbf{deploy SNR}& NA& NA& NA& 40dB& 35dB& 33dB& 31dB& 29dB& 22dB& 24dB& 48dB\\ 
 \hline
\end{tabular}
}
\end{table}

\label{sec:e2e_integration}

\section{Related-Work}


\textbf{Text-to-Video Diffusion Models}. 
Text-to-video (T2V) generation has rapidly advanced with diffusion-based architectures, which surpass GAN-based methods in temporal consistency and scalability. Early approaches extended text-to-image diffusion models by adding temporal layers to U-Net backbones, as seen in models like \emph{LTX-Video}~\cite{HaCohen2024LTXVideo} and \emph{Open-Sora Plan}~\cite{lin2024opensora}. However, these designs struggled with long-range temporal coherence and computational efficiency. Recent models increasingly adopt transformer-based architectures for their superior ability to model global spatio-temporal dependencies. For example, \emph{CogVideoX}~\cite{yang2024cogvideox} employs a diffusion transformer with a 3D VAE and expert transformer layers for strong text-video alignment. Similarly, \emph{Wan}~\cite{wan2025} adopts a large-scale transformer-based design for high-quality video synthesis, and \emph{HunYuan Video}~\cite{hunyuan2025} focuses on high-fidelity generation with joint image-video training and optimised text encoders.

Our work builds upon \emph{PyramidalFlow}~\cite{jin2024pyramidal}, which introduces a pyramidal flow matching strategy that progressively refines latents across spatial and temporal scales. Unlike cascaded pipelines, it unifies generation in a single diffusion transformer and supports autoregressive video generation with temporal pyramids. These inductive biases—hierarchical spatio-temporal modelling and autoregressive conditioning—make it a strong baseline for efficient, coherent video synthesis.\\

\noindent \textbf{On-device T2V Diffusion Models}.
Deploying text-to-video diffusion models on mobile devices introduces significant challenges due to limited compute and memory resources. Most existing mobile-optimised approaches build upon U-Net-based architectures, applying aggressive compression and pruning strategies:
AMD Hummingbird \cite{isobe2025amd} introduces a lightweight text-to-video generation framework that prunes large models by visual feedback learning to maintain quality while reducing parameters by 50\%. \emph{MobileVD}~\cite{benyahia2024mobilevd} adapts Stable Video Diffusion by reducing spatial resolution, introducing multi-scale temporal representations, and applying structured pruning, achieving over 500$\times$ efficiency gains with minimal quality degradation. Similarly, \emph{SnapGen-V}~\cite{wu2025snapgen} proposes a comprehensive acceleration framework that combines architecture search for temporal layers, adversarial fine-tuning, and step reduction.

While U-Net-based methods dominate existing mobile-optimised solutions, adapting transformer-based video diffusion models for on-device deployment remains an emerging and underexplored area. Recent parallel efforts have begun to investigate this direction by leveraging transformer-based denoisers. \emph{On-device Sora}~\cite{kim2025device} introduces a training-free adaptation of pre-trained diffusion models using linear proportional leap for reducing denoising steps, temporal token merging, and dynamic model loading to overcome memory constraints. \emph{Wu et al.}~\cite{wu2025taming} further pushes this direction by introducing a compressed VAE, KD-guided tri-level pruning, and adversarial step distillation, enabling video generation on mobile hardware. Most recently,~\cite{ghafoorian2025attention} introduced \emph{Attention Surgery}, a framework to distill pretrained state-of-the-art DiT models, into more efficient DiTs with hybrid self-attention.

\noindent \textbf{Text-encoder Distillation}.
Large text encoders, such as T5-XXL or CLIP, are widely used in diffusion-based generative models to capture rich semantic representations. However, their size and computational cost pose significant challenges for on-device deployment and real-time generation. To address this, \textit{DistillT5}~\cite{distillt5} introduces a vision-guided knowledge distillation framework that compresses large T5-based encoders into smaller variants (e.g., T5-Base) while preserving semantic alignment with the visual domain. The method employs multi-stage distillation using curated datasets optimised for image quality, semantic understanding, and text rendering, achieving up to 50$\times$ size reduction with minimal performance degradation. Related efforts in multimodal settings, such as \textit{CLIP distillation}~\cite{clipkd} and multilingual encoder distillation in \textit{AltDiffusion}~\cite{altdiffusion}, further demonstrate the effectiveness of encoder compression for improving efficiency in diffusion pipelines.

Within concurrent works on mobile-optimised T2V generation, text encoder optimisation remains largely overlooked. \textit{On-device Sora}~\cite{kim2025device} applies dynamic loading to T5 to reduce memory footprint but does not modify the encoder architecture itself. Similarly, \textit{Wu et al.}~\cite{wu2025taming} focuses on optimising the denoising backbone and VAE components, without introducing contributions towards text encoder compression. This highlights an open research gap in systematically distilling or compressing text encoders for efficient on-device video diffusion models that we attempt to fill with our novel contribution.\\

\noindent \textbf{Video Decoder Optimisation}.
While most research on efficiency in video diffusion models focuses on latent compression or denoising acceleration, decoder-side optimisation has received comparatively less attention. Existing works primarily explore architectural or inference-level strategies to reduce decoding overhead. \textit{LTX-Video}~\cite{HaCohen2024LTXVideo} introduces a decoder that performs the final denoising step, effectively shifting part of the refinement process from the diffusion backbone to the VAE decoder, reducing the number of diffusion iterations. \textit{WF-VAE} in Open-Sora Plan~\cite{lin2024opensora} proposes block-wise decoding with a \emph{Causal Cache} mechanism to enable tiled inference for high-resolution videos under memory constraints. Similarly, \textit{PyramidalFlow}~\cite{jin2024pyramidal} implements tile-enabled decoding and sequential offloading between CPU and GPU to support large-scale video generation on limited hardware. Cascaded approaches such as \textit{Imagen Video}~\cite{ho2022imagenvideo} and \textit{Lumiere}~\cite{lumiere2024} adopt multi-stage super-resolution decoders for progressive refinement, though these designs prioritise quality over on-device efficiency.

In contrast, our work introduces an \textit{asymmetric decoder distillation} strategy that substitutes the base model’s native decoder with a device-optimised architecture while preserving the original video encoding scheme. 
Unlike prior methods that rely on tiling or caching for memory savings, our method directly targets decoder complexity through knowledge distillation, enabling efficient deployment without altering the latent representation or retraining the diffusion backbone.\\

\noindent \textbf{Block Pruning}. 
Pruning for diffusion transformers has recently gained attention as a means to reduce inference cost without retraining from scratch. \emph{TinyFusion}~\cite{fang2025tinyfusion} introduces a learnable depth-pruning framework for DiTs, where layer masks are optimised jointly with a recoverability objective and refined through masked knowledge distillation. Similarly, \emph{Effortless Efficiency}~\cite{zhang2025effortless} proposes a model-agnostic structural pruning approach for diffusion models, learning pruning masks across the denoising process to remove redundant layers with minimal fine-tuning. For video diffusion transformers, the parallel work by \emph{Wu et~al.}~\cite{wu2025taming} adopts a sensitivity-aware tri-level pruning strategy that prunes at multiple granularities—within-layer components, attention heads, and entire blocks—guided by knowledge distillation and sensitivity analysis, as part of a broader system-level optimisation for real-time mobile generation.

In contrast, our method focuses on \emph{block-level pruning} tailored to the MMDiT denoiser. We introduce a three-stage pipeline comprising block-importance scoring, short fine-tuning, and full teacher-model distillation. Unlike TinyFusion’s differentiable depth pruning or Wu et~al.’s multi-granularity sensitivity-based approach, our strategy aligns pruning units with the natural MMDiT block structure to preserve spatio-temporal attention pathways critical for video generation, while simplifying the pruning process for practical deployment.\\

\noindent \textbf{Step Distillation}.
Reducing the number of denoising steps in diffusion models is critical for improving inference efficiency, and several step distillation strategies have been proposed. \emph{Progressive Distillation}~\cite{salimans2022progressive} is a seminal approach that iteratively halves the number of steps by training a student to mimic the teacher’s trajectory, achieving substantial speedups while preserving quality. Subsequent works explore alternative paradigms, such as \emph{adversarial step distillation}~\cite{zhang2024sf}, which augments the distillation objective with adversarial losses to enhance perceptual fidelity; this strategy has been adopted in video generation pipelines such as ~\cite{HaCohen2024LTXVideo, wu2025taming}. Another interesting direction is \emph{Distribution Matching Distillation (DMD)}~\cite{salimans2022progressive, song2023consistency}, which aligns the student’s output distribution with that of the teacher across noise levels, providing a principled framework for step reduction without progressive halving. Building on this foundation, we are the first to adapt a DMD-based step distillation method to a pyramidal flow-matching denoiser~\cite{jin2024pyramidal}, enabling efficient inference while preserving the model’s hierarchical spatio-temporal structure.
\section{Conclusion}
In this work, we introduced four key optimisations—Text-Encoder Distillation, Asymmetric Decoder Distillation, Block Pruning, and Step Distillation—that collectively transform the \baseline~\cite{jin2024pyramidal} pipeline into \name, an efficient, on-device text-to-video generation system. These innovations reduce latency from minutes to seconds while preserving state-of-the-art quality, marking a significant step toward practical, interactive video generation on consumer hardware. Yet, this achievement is not an endpoint but a beginning. Our GPT-pretraining-scale text-to-video model serves as a foundational vehicle for building a new class of real-time, interactive generative editing applications. The holy grails of this domain—long-form video generation and true real-time synthesis—remain open frontiers, demanding continued exploration. Looking ahead, we envision several promising directions: (i) leveraging stronger foundation models such as WAN~\cite{wan2025}, (ii) incorporating more efficient transformer alternatives such as linear/hybrid attention~\cite{ghafoorian2025attention,chen2025sana} and token pruning~\cite{kahatapitiya2024object,peruzzo2025adaptor}, (iii) adopting SSM-based VAEs for efficient and expressive video tokenisation, (iv) enabling video-conditioned on-device generation, (v) exploring recurrent architectures for temporal coherence, (vi) incorporating more advanced and compressive auto-encoders~\cite{chen2025dc} and better representing the motion~\cite{bhowmik2025moalign}. Just as the Ship of Theseus invites us to reflect on continuity and change, \name~embodies the evolving identity of generative systems—where each optimisation replaces a part, yet the spirit of the original persists. In this evolution lies the future of generative video: adaptive, modular, and ultimately capable of delivering seamless, real-time creative experiences.


\end{document}